\documentclass[english]{article}

\newcommand\accepted

\ifdefined\preprint
\usepackage[preprint]{neurips_2019}
\else
\ifdefined\accepted
\usepackage[final]{neurips_2019}
\else
\usepackage{neurips_2019}
\fi
\fi

\usepackage[T1]{fontenc}
\usepackage[utf8]{inputenc}
\PassOptionsToPackage{obeyFinal}{todonotes}
\usepackage{xcolor}
\usepackage[english]{babel}
\usepackage{mathtools}
\usepackage{url}
\usepackage{todonotes}
\usepackage{amsmath}
\usepackage{amssymb}
\usepackage{stmaryrd}
\usepackage{graphicx}
\usepackage{booktabs}
\usepackage{hyperref}
\usepackage{multirow}
\usepackage{natbib}
\usepackage{microtype}
\usepackage{subfig}
\usepackage{amsfonts}
\usepackage{nicefrac}
\usepackage{longtable}
\usepackage{appendix}
\usepackage{caption}
\usepackage{wrapfig}
\usepackage{floatrow}
\usepackage[ruled, vlined, linesnumbered]{algorithm2e}

\title{Unsupervised Scalable Representation Learning\\ for Multivariate Time Series}

\author{%
  Jean-Yves Franceschi\thanks{Work partially done while studying at ENS de Lyon and MLO, EPFL.} \\
  Sorbonne Université, CNRS, LIP6, F-75005 Paris, France \\
  \texttt{jean-yves.franceschi@lip6.fr} \\
  \And
  Aymeric Dieuleveut \\
  MLO, EPFL, Lausanne CH-1015, Switzerland \\
  CMAP, Ecole Polytechnique, Palaiseau, France \\
  \texttt{aymeric.dieuleveut@polytechnique.edu} \\
  \And
  Martin Jaggi \\
  MLO, EPFL, Lausanne CH-1015, Switzerland \\
	\texttt{martin.jaggi@epfl.ch}
}

\definecolor{mydarkblue}{rgb}{0,0.08,0.45}
\hypersetup{ %
  pdftitle={},
  pdfauthor={},
  pdfsubject={},
  pdfkeywords={},
  pdfborder=0 0 0,
  pdfpagemode=UseNone,
  colorlinks=true,
  linkcolor=mydarkblue,
  citecolor=mydarkblue,
  filecolor=mydarkblue,
  urlcolor=mydarkblue,
  pdfview=FitH}

\makeatletter
\providecommand{\tabularnewline}{\\}
\makeatother

  \providecommand{\ff}{\boldsymbol{f}}

  \providecommand{\xx}{\boldsymbol{x}}
  \providecommand{\yy}{\boldsymbol{y}}
  
  \providecommand{\ttheta}{\boldsymbol{\theta}}

\begin{document}


\maketitle

\begin{abstract}
Time series constitute a challenging data type for machine learning
algorithms, due to their highly variable lengths and sparse labeling in practice.
In this paper, we tackle this challenge by proposing an unsupervised method to learn
universal embeddings of time series. Unlike previous works,
it is scalable with respect to their length and we demonstrate
the quality, transferability and practicability of the learned representations
with thorough experiments and comparisons. To this end, we combine an encoder
based on causal dilated convolutions with a novel triplet loss employing time-based
negative sampling, obtaining general-purpose representations for variable length and multivariate time series.
\end{abstract}

\global\long\def\size{\operatorname{size}}
\DeclarePairedDelimiter\parentheses{(}{)}
\DeclarePairedDelimiter\braces{\{}{\}}
\DeclarePairedDelimiter\brackets{[}{]}
\DeclarePairedDelimiter\lbracket{[}{.}
\DeclarePairedDelimiter\rbracket{.}{]}
\DeclarePairedDelimiter\lrbrackets{\llbracket}{\rrbracket}
\DeclarePairedDelimiter\euclideannorm{\|}{\|}

\section{Introduction}

We investigate in this work the topic of unsupervised
general-purpose representation learning for time series. In spite
of the increasing amount of work about representation learning in
fields like natural language processing \citep{Young2018} or videos \citep{Denton2017}, few articles explicitly
deal with general-purpose representation learning for time series without structural assumption on
non-temporal data.

This problem is indeed challenging for various reasons.
First, real-life time series are rarely or sparsely labeled. Therefore,
\emph{unsupervised} representation learning would be strongly preferred.
Second, methods need to deliver compatible representations while allowing the input time series to have unequal lengths. Third, scalability and efficiency
both at training and inference time is crucial, in the sense that the techniques must work for both short and long time series encountered in practice.

Hence, we propose in the following an \emph{unsupervised} method to learn \emph{general-purpose representations} for \emph{multivariate}
time series that comply with the issues of \emph{varying and potentially
high lengths} of the studied time series. To this end, we introduce a novel
unsupervised loss training a scalable encoder, shaped as a deep convolutional
neural network with dilated convolutions \citep{Oord2016} and outputting fixed-length
vector representations regardless of the length of its output. This loss is built
as a triplet loss employing time-based negative sampling, taking advantage of the
encoder resilience to time series of unequal lengths. To our knowledge, it is the
first fully unsupervised triplet loss in the literature of time series.

We assess the quality of the learned representations on various datasets
to ensure their universality. In particular, we test how our representations
can be used for classification tasks on the standard datasets in the time series
literature, compiled in the UCR repository \citep{Dau2018}.
We show that our representations are \emph{general}
and \emph{transferable}, and that our method \emph{outperforms concurrent
unsupervised methods} and even \emph{matches the state of the art} of
non-ensemble supervised classification techniques.
Moreover, since UCR time series are exclusively univariate and mostly short,
we also evaluate our representations on the
recent UEA multivariate time series repository \citep{Bagnall2018},
as well as on a real-life dataset including very long time series,
on which we demonstrate \emph{scalability}, \emph{performance} and generalization ability \emph{across different tasks} beyond classification.

This paper is organized as follows. Section~\ref{sec:RelatedWork}
outlines previous works on unsupervised representation
learning, triplet losses and deep architectures for time series in the literature. Section~\ref{sec:Training}
describes the unsupervised training of the encoder, while Section~\ref{sec:EncoderArchitecture}
details the architecture of the latter. Finally, Section~\ref{sec:ExperimentalResults}
provides results of the experiments that we conducted to evaluate
our method.

\section{\label{sec:RelatedWork}Related Work}

\paragraph*{Unsupervised learning for time series.}
To our knowledge, apart from those dealing with videos or high-dimensional data
\citep{Srivastava2015,Denton2017,Villegas2017,Oord2018},
few recent works tackle unsupervised representation learning
for time series.
\citet{Fortuin2019} deal with a related but different problem to this work,
by learning temporal representations of time series that represent well their
evolution.
\citet{Hyvarinen2016} learn representations on evenly sized subdivisions of
time series by learning to discriminate between those subdivisions from these
representations. \citet{Lei2017} expose an unsupervised method designed so that the
distances between learned representations mimic a standard distance (Dynamic
Time Warping, DTW) between time series. \citet{Malhotra2017} design an encoder
as a recurrent neural network, jointly trained with a decoder as a
sequence-to-sequence model to reconstruct the input time series from its
learned representation. Finally, \citet{Wu2018} compute feature embeddings
generated in the approximation of a carefully designed and efficient kernel.

However, these methods either are not scalable nor suited to long time series
(due to the sequential nature of a recurrent network, or to the use of DTW with
a quadratic complexity with respect to the input length), are tested on no or
very few standard datasets and with no publicly available code, or do not provide sufficient comparison
to assess the quality of the learned representations. Our scalable model and
extensive analysis aim at overcoming these issues, besides outperforming these methods.

\paragraph*{Triplet losses.}
Triplet losses have recently been widely used in various forms for representation learning in
different domains \citep{Mikolov2013, Schroff2015, Wu2018Starspace} and
have also been theoretically studied \citep{Arora2019}, but have not
found much use for time series apart from audio
\citep{Bredin2017, Lu2017b, Jansen2018}, and never, to our
knowledge, in a fully unsupervised
setting, as existing works assume the existence of class labels or annotations in the
training data. Closer to our work even though focusing on a different, more specific task,
\citet{Turpault2019} learn audio embeddings in a semi-supervised setting, while partially relying
on specific transformations of the training data to sample positive samples in
the triplet loss; \citet{Logeswaran2018} train a sentence encoder
to recognize, among randomly chosen sentences, the true context of another sentence,
which is a difficult method to adapt to time series. Our method instead relies on a more
natural choice of positive samples, learning similarities using subsampling.

\paragraph*{Convolutional networks for time series.}
Deep convolutional neural networks have recently been successfully applied to time series classification tasks \citep{Cui2016,Wang2017}, showing competitive
performance. Dilated convolutions, popularized by WaveNet \citep{Oord2016} for
audio generation, have been used to improve their performance and were shown
to perform well as sequence-to-sequence models for time series forecasting
\citep{Bai2018} using an architecture that inspired ours. These works
particularly show that dilated convolutions help
to build networks for sequential tasks outperforming recurrent
neural networks in terms of both efficiency and prediction performance.

\section{\label{sec:Training}Unsupervised Training}

We seek to train an encoder-only architecture, avoiding the need to jointly train with a decoder as in autoencoder-based standard representation learning methods as done by \citet{Malhotra2017}, since those would induce a larger computational cost.
To this end, we introduce a novel triplet loss for time series, inspired by the successful and by now classic word representation learning method known as word2vec \citep{Mikolov2013}.
The proposed triplet loss uses original time-based sampling strategies to overcome the
challenge of learning on unlabeled data. As far as we know, this work is the
first in the time series literature to rely on a triplet loss in a fully
unsupervised setting.

\begin{wrapfigure}{R}{0.5\columnwidth}
\resizebox{\columnwidth}{!}{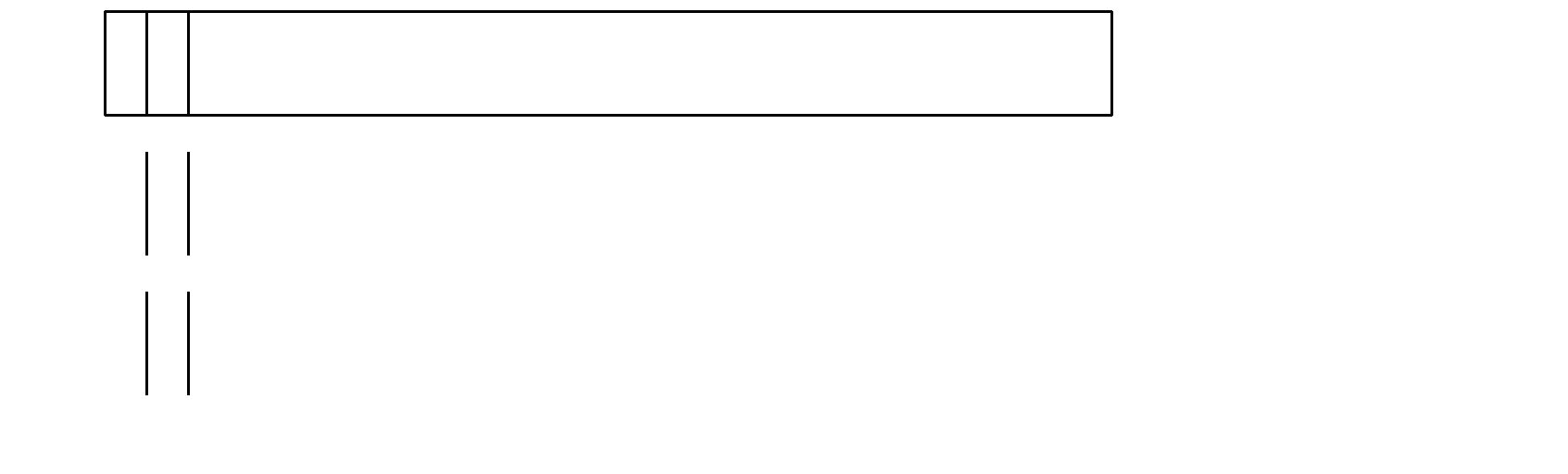}
\caption{\label{fig:TrainingIllustration}Choices of $\xx^{\mathrm{ref}}$, $\xx^{\mathrm{pos}}$ and $\xx^{\mathrm{neg}}$.}
\end{wrapfigure}

The objective is to ensure that similar time series
obtain similar representations, with no supervision to learn such similarity.
Triplet losses help to achieve the former \citep{Schroff2015}, but require to
provide pairs of similar inputs, thus challenging the latter.
While previous supervised works for time series using triplet losses assume that
data is annotated, we introduce an
unsupervised time-based criterion to select pairs of similar time series and
taking into account time series of varying lengths, by following word2vec's intuition.
The assumption made in the CBOW model of word2vec is twofold.
The representation of the \emph{context} of a word should probably be, on one hand,
close to the one of this word \citep{Goldberg2014}, and, on the other hand, distant from the one of randomly
chosen words, since they are probably unrelated to the original word's
context. The corresponding loss then pushes pairs of (context, word) and (context, random word)
to be linearly separable. This is called \emph{negative sampling}.

To adapt this principle to time series, we consider (see Figure~\ref{fig:TrainingIllustration} for an
illustration) a random
subseries\footnote{I.e., a subsequence of a time series composed by consecutive time
steps of this time series.}
$\xx^{\mathrm{ref}}$ of a given time series~$\yy_i$. Then, on one hand, the representation of $\xx^{\mathrm{ref}}$
should be close to the one of any of its subseries $\xx^{\mathrm{pos}}$
(a \emph{positive} example). On the other
hand, if we consider another subseries $\xx^{\mathrm{neg}}$ (a \emph{negative}
example) chosen at random (in a different random time series $\yy_j$ if several series are available, or in the same time series if it is long enough and not stationary), then
its representation should be distant from
the one of $\xx^{\mathrm{ref}}$. Following the analogy with word2vec,
$\xx^{\mathrm{pos}}$ corresponds to a word, $\xx^{\mathrm{ref}}$ to its context,
and $\xx^{\mathrm{neg}}$ to a random word.
To improve the stability and convergence of the training procedure
as well as the experimental results of our learned representations,
we introduce, as in word2vec, several
negative samples $\parentheses*{\xx_{k}^{\mathrm{neg}}}_{k\in \lrbrackets*{1,K}}$,
chosen independently at random.

The objective corresponding to these choices to minimize during training can be thought of the one of word2vec with its shallow network replaced by a deep network
$\ff\!\left(\cdot,\ttheta\right)$ with parameters $\ttheta$, or formally
\begin{equation}
	\label{eq:def_loss}
	- \log\parentheses*{\sigma\parentheses*{\ff\parentheses*{\xx^{\mathrm{ref}},\ttheta}^\top \ff\parentheses*{\xx^{\mathrm{pos}},\ttheta}}}
	- \sum_{k=1}^{K}\log\parentheses*{\sigma\parentheses*{-\ff\parentheses*{\xx^{\mathrm{ref}},\ttheta}^\top \ff\parentheses*{\xx_{k}^{\mathrm{neg}},\ttheta}}},
\end{equation}
where $\sigma$ is the sigmoid function. This loss pushes the
computed representations to distinguish between $\xx^{\mathrm{ref}}$
and $\xx^{\mathrm{neg}}$, and to assimilate $\xx^{\mathrm{ref}}$ and
$\xx^{\mathrm{pos}}$.
Overall, the training procedure consists in traveling through the
training dataset for several epochs (possibly using mini-batches),
picking tuples $\parentheses*{\xx^{\mathrm{ref}},\xx^{\mathrm{pos}},\parentheses*{\xx^{\mathrm{neg}}_k}_k}$
at random as detailed in Algorithm~\ref{alg:TripletChoice}, and performing
a minimization step on the corresponding loss
for each pair, until training ends. The overall
computational and memory cost is $\mathcal{O}\parentheses*{K \cdot c\parentheses*{\ff}}$,
where $c\parentheses*{\ff}$ is the cost of evaluating and backpropagating through $\ff$
on a time series; thus this unsupervised training is
scalable as long as the encoder architecture is scalable as well.

The length of the negative examples is chosen at random in Algorithm~\ref{alg:TripletChoice}
for the most general case; however, their length can also be the same for all
samples and equal to $\size\parentheses*{\xx^{\mathrm{pos}}}$.
The latter case is suitable when all time series in the dataset have
equal lengths, and speeds up the training procedure thanks to
computation factorizations; the former case is only used when time
series in the dataset do not have the same lengths, as we experimentally saw no other
difference than time efficiency between the two cases.
In our experiments, we do not cap the lengths of $\xx^{\mathrm{ref}}$, $\xx^{\mathrm{pos}}$ and $\xx^{\mathrm{neg}}$
since they are already limited by the length of the train time series, which corresponds
to scales of lengths on which our representations are tested.

We highlight that this time-based triplet loss leverages the ability of
the chosen encoder to take as input time series of different lengths. By
training the encoder on a range of input lengths going from one to the length
of the longest time series in the train set, it becomes able to output
meaningful and transferable representations regardless of the input length, as shown in
Section~\ref{sec:ExperimentalResults}.

This training procedure is interesting in that it is efficient enough
to be run over long time series (see Section~\ref{sec:ExperimentalResults})
with a scalable encoder (see Section~\ref{sec:EncoderArchitecture}), thanks to
its decoder-less design and the separability of the loss, on which
a backpropagation per term can be performed to save memory.\footnote{
We used this optimization for multivariate or long (with length
higher than $10\,000$) time series.
}

\IncMargin{1em}
\begin{algorithm}[t]
	\For{$i\in \lrbrackets*{1,N}$ \textbf{\textup{with}} $s_i = \size\parentheses*{\yy_i}$}{
		pick $s^{\mathrm{pos}} = \size\parentheses*{\xx^{\mathrm{pos}}}$ in $\lrbrackets*{1,s_i}$ and $s^{\mathrm{ref}} = \size\parentheses*{\xx^{\mathrm{ref}}}$ in $\lrbrackets*{s^{\mathrm{pos}},s_i}$ uniformly at random\;
		pick $\xx^{\mathrm{ref}}$ uniformly at random among subseries of $\yy_i$ of length $s^{\mathrm{ref}}$\;
		pick $\xx^{\mathrm{pos}}$ uniformly at random among subseries of $\xx^{\mathrm{ref}}$ of length $s^{\mathrm{pos}}$\;
		\DontPrintSemicolon
		pick uniformly at random $i_k \in \lrbrackets*{1,N}$, then $s_k^{\mathrm{neg}}=\size\parentheses*{\xx^{\mathrm{neg}}_k}$ in $\lrbrackets{1,\size\parentheses*{\yy_k}}$ and finally $\xx^{\mathrm{neg}}_k$ among subseries of $\yy_k$ of length $s_k^{\mathrm{neg}}$, for $k \in \lrbrackets*{1,K}$.\;
	}
	\caption{\label{alg:TripletChoice}Choices of $\xx^{\mathrm{ref}}$, $\xx^{\mathrm{pos}}$ and $\parentheses*{\xx^{\mathrm{neg}}_k}_{k \in \lrbrackets*{1,K}}$ for an epoch over the set $\parentheses*{\yy_i}_{i\in \lrbrackets*{1,N}}$.}
\end{algorithm}
\DecMargin{1em}

\section{\label{sec:EncoderArchitecture}Encoder Architecture}

We explain and present in this section our choice of architecture for the encoder, which is motivated by
three requirements: it must extract relevant information from time series;
it needs to be time- and memory-efficient, both for training and testing;
and it has to allow variable-length inputs.
We choose to use deep neural networks with \emph{exponentially dilated causal convolutions}
to handle time series. While they have been popularized in the context of
sequence generation \citep{Oord2016}, they have never been used for unsupervised time series
representation learning. They offer several advantages.

Compared to recurrent neural
networks, which are inherently designed for sequence-modeling tasks and
thus sequential, these networks are scalable as they allow efficient parallelization
on modern hardware such as GPUs. Besides this demonstrated efficiency,
exponentially dilated convolutions have also been introduced
to better capture, compared to full convolutions, long-range dependencies
at constant depth by exponentially increasing the receptive
field of the network \citep{Oord2016,Yu2016,Bai2018}.

Convolutional networks have also been demonstrated to be performant on various
aspects for sequential data.
For instance, recurrent networks are known to be subject to the issue of exploding
and vanishing gradients, due to their recurrent nature \citep[Chapter~10.9]{Goodfellow2016}.
While significant work has been done to tackle this issue
and improve their ability to capture long-term dependencies, such as the
LSTM \citep{Hochreiter1997}, recurrent networks are still outperformed
by convolutional networks on this aspect \citep{Bai2018}.
On the specific domains of time series classification, which is an
essential part of our experimental evaluation, and forecasting, deep neural networks
have recently been successfully used \citep{Bai2018,IsmailFawaz2019}.

\begin{figure}
\begin{centering}
\subfloat[\label{fig:CausalCNNPrinciple}]{\begin{centering}
\resizebox{0.6\columnwidth}{!}{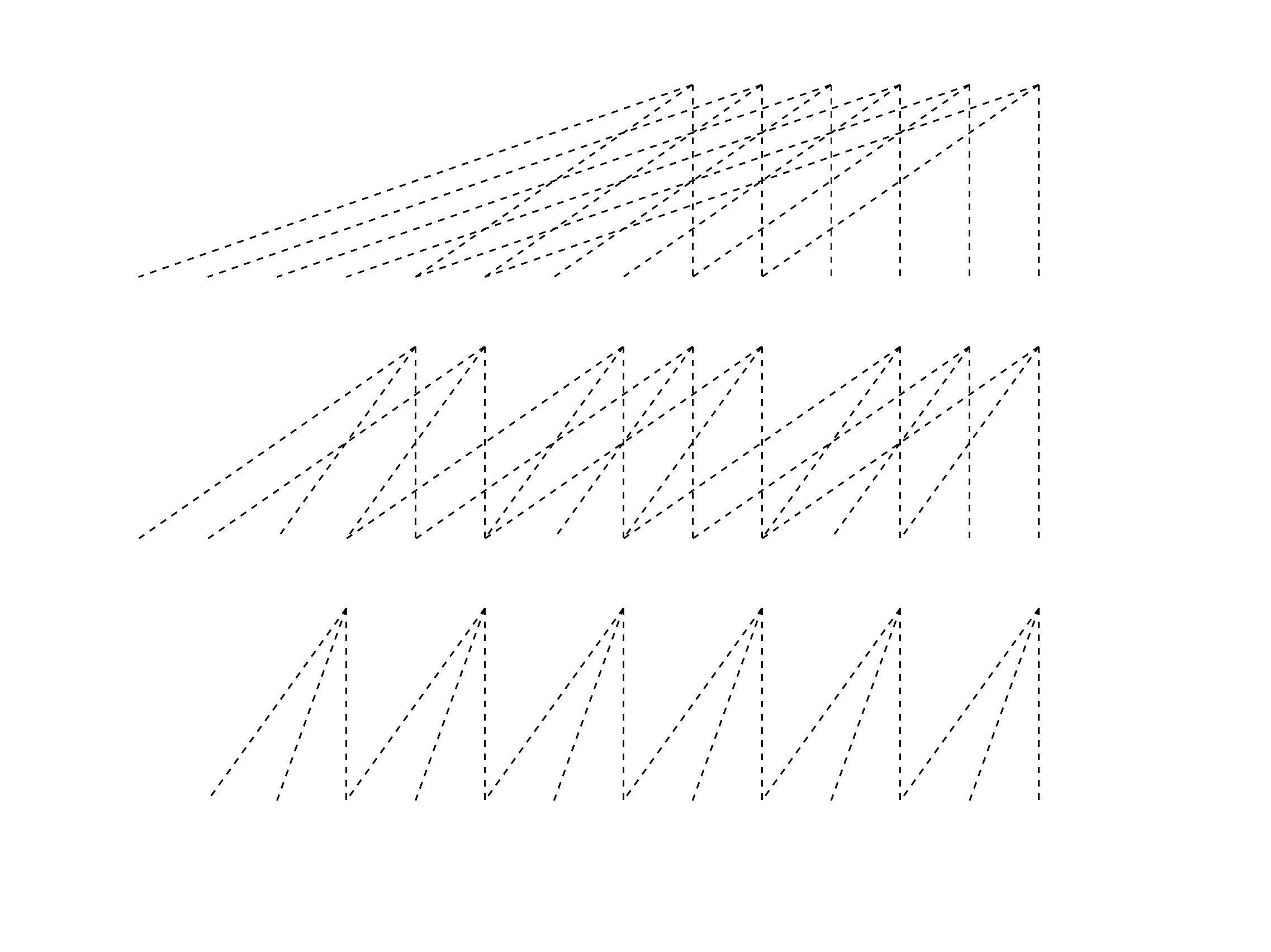}
\par\end{centering}
\centering{}}
\subfloat[\label{fig:Layer}]{\begin{centering}
\resizebox{0.355\columnwidth}{!}{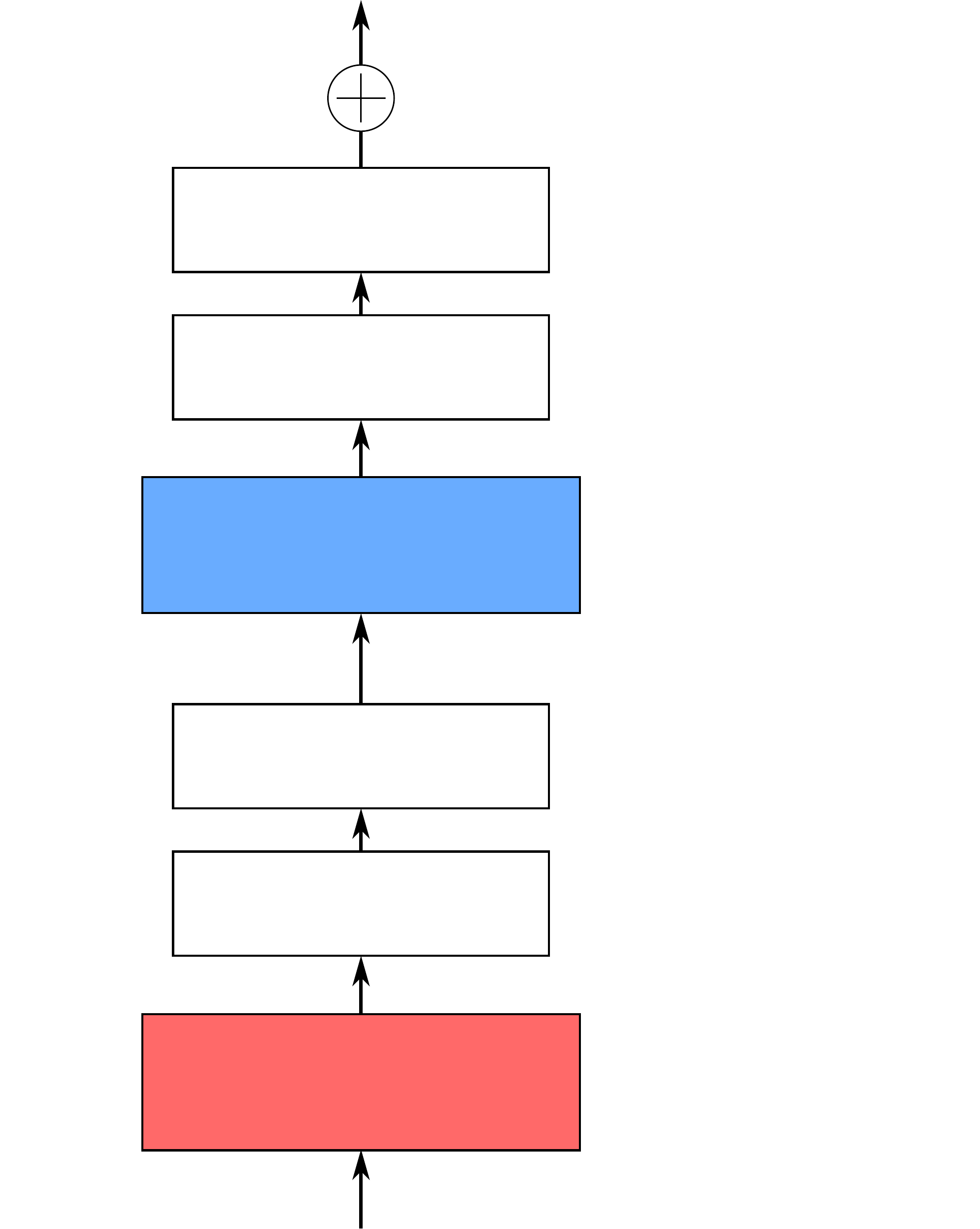}
\par\end{centering}
}
\par\end{centering}
\caption{(a) Illustration of three stacked dilated causal
convolutions. Lines between each sequence represent their
computational graph.
Red solid lines highlight the dependency graph for the computation
of the last value of the output sequence, showing that no future
value of the input time series is used to compute it. (b) Composition of the
$i$-th layer of the chosen architecture.}
\end{figure}

Our model is particularly based on stacks of dilated \emph{causal} convolutions
(see Figure~\ref{fig:CausalCNNPrinciple}), which map a sequence to a sequence of the same length,
such that the $i$-th element of the output sequence is computed using
only values up until the $i$-th element of the input sequence, for
all $i$. It is thus called causal, since the output value
corresponding to a given time step is not computed using future input
values.
Causal convolutions allow alleviating the disadvantage of not
using recurrent networks at testing time. Indeed, recurrent networks
can be used in an online fashion, thus saving memory and computation time during
testing. In our case, causal convolutions organize the computational graph so that,
in order to update its output when an element is added at the end of the input time series,
one only has to evaluate the highlighted graph shown in Figure~\ref{fig:CausalCNNPrinciple}
rather than the full graph.

Inspired by \citet{Bai2018}, we build each layer
of our network as a combination of causal convolutions, weight
normalizations \citep{Salimans2016}, leaky ReLUs and residual connections
(see Figure~\ref{fig:Layer}). Each of these layers is given an exponentially
increasing dilation parameter ($2^{i}$ for the $i$-th layer).
The output of this causal network is then given to
a global max pooling layer squeezing the temporal dimension
and aggregating all temporal information in a fixed-size vector (as
proposed by \citet{Wang2017} in a supervised setting with full convolutions).
A linear transformation of this vector is then the output of the encoder,
with a fixed, independent from the input length, size.

\section{\label{sec:ExperimentalResults}Experimental Results}

We review in this section experiments conducted to investigate the relevance
of the learned representations.
\ifdefined\preprint
The code corresponding to these experiments is publicly available.\footnote{\url{https://github.com/White-Link/UnsupervisedScalableRepresentationLearningTimeSeries}.}
\else
\ifdefined\accepted
The code corresponding to these experiments is
attached in the supplementary material and is publicly available.\footnote{\url{https://github.com/White-Link/UnsupervisedScalableRepresentationLearningTimeSeries}.}
\else
The code corresponding to these experiments is
attached in the supplementary material and will be publicly released upon
publication.
\fi
\fi
The full training process and hyperparameter choices are detailed in the
supplementary material, Sections~\ref{app:detailed_training} and~\ref{app:hyper_params}.
We used Python 3 for implementation, with PyTorch 0.4.1 \citep{Paszke2017}
for neural networks and scikit-learn \citep{Pedregosa2011} for SVMs.
Each encoder was trained using the Adam optimizer \citep{Kingma2015} on a single Nvidia Titan Xp GPU with CUDA
9.0, unless stated otherwise.

Selecting hyperparameters for an unsupervised method is challenging since the plurality of
downstream tasks is usually supervised. Therefore, as \cite{Wu2018}, we choose for each
considered dataset archive a single set of hyperparameters regardless of the downstream task.
Moreover, we highlight that we perform \emph{no hyperparameter optimization} of the
unsupervised encoder architecture and training parameters for any task,
unlike other unsupervised works such as TimeNet \citep{Malhotra2017}.
Particularly, for classification tasks, \emph{no label} was used during the encoder training.

\subsection{Classification}

We first assess the \emph{quality} of our learned representations on supervised
tasks in a standard manner \citep{Xu2003, Ghahramani2014} by using
them for time series classification. In this setting, we show that
(1) our method outperforms state-of-the-art unsupervised methods, and
notably achieves performance close to the supervised state of the art,
(2) strongly outperforms supervised deep learning methods when data is
only sparsely labeled, (3) produces transferable representations.

For each considered dataset with a train~/~test split, we unsupervisedly
train an encoder using its train set. We then train an SVM with radial
basis function kernel on top of the
learned features using the train labels of the dataset, and
output the corresponding classification score on the test set. As
our training procedure encourages representations of different time
series to be separable, observing the classification performance of
a simple SVM on these features allows to check their quality \citep{Wu2018}.
Using SVMs also allows, when the encoder is trained, an
\emph{efficient} training both in terms of time (training is a matter of minutes in most cases) and
space.

As $K$ has a significant impact on the performance, we
present a \emph{combined} version of our method,
where representations computed by encoders trained with different values of $K$
(see Section~\ref{app:hyper_params} for more details) are concatenated.
This enables our learned representations
with different parameters to complement each other, and to remove
some noise in the classification scores.

\subsubsection{Univariate Time Series}

We present accuracy scores for all 128 datasets of the new iteration of the UCR archive \citep{Dau2018},
which is a standard set of varied univariate datasets.
We report in Table~\ref{tab:Ours}
scores for only some UCR datasets, while scores for all datasets are
reported in the supplementary material, Section~\ref{app:full_univariate_results}.

We first compare our scores to the two concurrent methods of this work,
TimeNet \citep{Malhotra2017} and RWS \citep{Wu2018}, which are two unsupervised methods
also training a simple classifier on top of the learned representations,
and reporting their results on a few UCR datasets.
We also compare on the first 85 datasets of the
archive\footnote{The new UCR archive includes 43 new datasets on which no reproducible
results of state-of-the-art methods have been produced yet. Still, we provide complete results for our method on these datasets in the supplementary material, Section~\ref{app:full_univariate_results}, Table~\ref{app_table:our_results_ucr_newest}, along with those of DTW, the only other method for which they were available.}
to the four best classifiers of the supervised state of the art
studied by \citet{Bagnall2017}:
COTE -- replaced by its improved version HIVE-COTE \citep{Lines2018} --,
ST \citep{Bostrom2015}, BOSS \citep{Schaefer2015} and EE \citep{Lines2015}.
HIVE-COTE is a powerful ensemble method using many
classifiers in a hierarchical
voting structure; EE is a simpler ensemble method;
ST is based on shapelets and BOSS is a dictionary-based
classifier.\footnote{While ST and BOSS are also ensembles of classifiers,
we chose not to qualify both of them as ensembles since their ensemble only
includes variations of the same novel classification method.}
We also add DTW (one-nearest-neighbor classifier with DTW as measure) as a baseline.
HIVE-COTE includes ST, BOSS, EE and DTW in
its ensemble, and is thus expected to outperform them.
Additionally, we compare our method to the ResNet method of \citet{Wang2017}, which is
the best supervised neural network method studied in the review of \citet{IsmailFawaz2019}.

\begin{figure}
\begin{centering}
\begin{floatrow}
\ffigbox{
	\includegraphics[width=\linewidth]{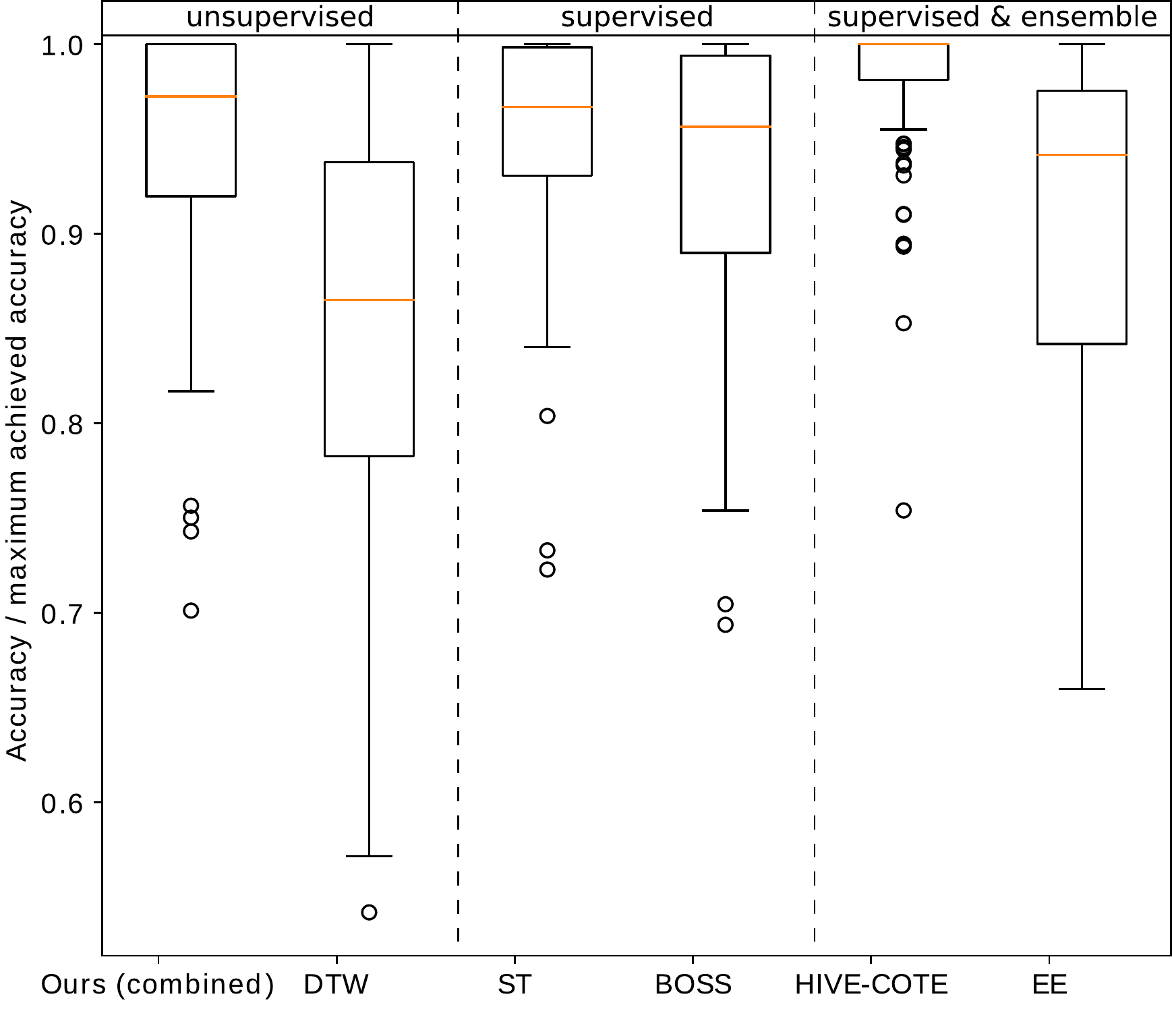}
}{\caption{\label{fig:BoxPlot}Boxplot of the ratio of the accuracy
versus maximum achieved accuracy (higher is better) for compared methods on the first 85 UCR datasets.}}
\ffigbox{
	\includegraphics[width=\linewidth]{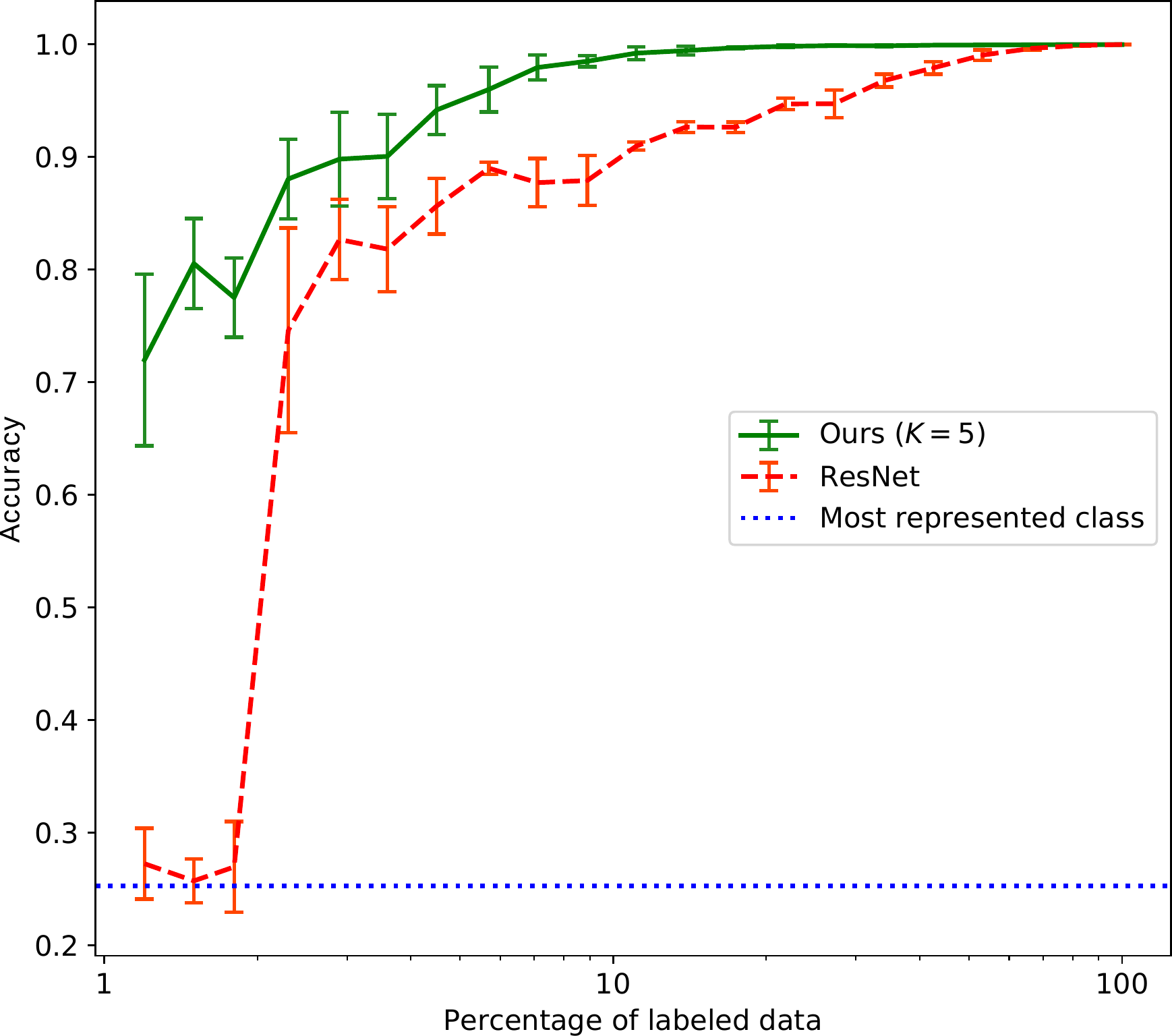}
}{\caption{\label{fig:SparseLabeling}Accuracy of ResNet and our method with respect to the ratio of labeled data on TwoPatterns. Error bars correspond to the standard deviation over five runs per point for each method.}}
\end{floatrow}
\end{centering}
\end{figure}

\paragraph*{Performance.}

Comparison with the unsupervised state of the art (Section~\ref{app:full_univariate_results}, Table~\ref{app_table:comparison_to_others_deep} of the supplementary material), indicates that our method
\emph{consistently matches or outperforms both unsupervised methods}
TimeNet and RWS (on 11 out of 12 and 10 out of
11 UCR datasets), showing its performance. Unlike our work, code and full
results on the UCR archive are not provided for these methods, hence the incomplete results.

When comparing to the supervised non-neural-network state of the art, we observe (see
Figures~\ref{fig:CriticalDifferenceDiagram} and~\ref{fig:Histogram} in the supplementary material)
that our method is globally the second-best one (with average rank $2.92$),
only beaten by HIVE-COTE ($1.71$) and equivalent to ST ($2.95$).
Thus, our unsupervised method beats several
recognized supervised classifier, and is only preceded by a powerful
ensemble method, which was expected since the latter takes advantage of
numerous classifiers and data representations. Additionally, Figure~\ref{fig:BoxPlot}
shows that our method has the second-best median for
the ratio of accuracy over maximum achieved accuracy, behind HIVE-COTE
and above ST.
Finally, results reported from the study of \citet{IsmailFawaz2019} for the fully supervised ResNet (Section~\ref{app:full_univariate_results}, Table~\ref{app_table:comparison_to_others_deep} of the supplementary material) show that it expectedly outperforms our method on $63\%$
out of 71 UCR datasets.\footnote{Those results are incomplete as \citet{IsmailFawaz2019} performed their experiments on
the old version of the archive, whereas ours are performed on its most recent release where
some datasets were changed.}
Overall, our method achieves remarkable performance as it \emph{is close to the best supervised neural network},
\emph{matches the second-best studied non-neural-network supervised method}, and, in particular,
is \emph{at the level of the best performing method included in HIVE-COTE}.\footnote{Our method could be included in HIVE-COTE, which could improve its performance, but this is beyond the scope of this work
and requires technical work, as HIVE-COTE is implemented in Java and ours in Python.}

\begin{table}
\caption{\label{tab:Ours}Accuracy scores of variants of our method compared with other supervised and unsupervised methods, on some
UCR datasets. Results for the whole archive are available in the supplementary material, Section~\ref{app:full_univariate_results}, Tables~\ref{app_table:our_results_ucr},~\ref{app_table:compar_results_ucr} and~\ref{app_table:our_results_ucr_newest}. Bold and underlined scores respectively indicate the best and second-best (when there is no tie for first place) performing methods.}
\begin{centering}
\scriptsize
\begin{tabular}{lccccccccccc}
\toprule
\multirow{3}{*}[-\dimexpr \aboverulesep + \belowrulesep + \cmidrulewidth]{{Dataset}} & \multicolumn{5}{c}{{Unsupervised}} & \multicolumn{4}{c}{{Supervised}} \tabularnewline
\cmidrule(lr){2-6} \cmidrule(lr){7-10}
& \multicolumn{4}{c}{{Ours}} & \multirow{2}{*}[-0.5\dimexpr \aboverulesep + \belowrulesep + \cmidrulewidth]{{DTW}} & \multirow{2}{*}[-0.5\dimexpr \aboverulesep + \belowrulesep + \cmidrulewidth]{{ST}} & \multirow{2}{*}[-0.5\dimexpr \aboverulesep + \belowrulesep + \cmidrulewidth]{{BOSS}} & \multicolumn{2}{c}{{Ensemble}} \tabularnewline
\cmidrule(lr){2-5} \cmidrule(lr){9-10}
& {$K=5$} & {$K=10$} & {Combined} & {FordA} & & & & {HIVE-COTE} & {EE} \tabularnewline
\midrule
{DiatomSizeReduction} & \textbf{0.993} & {0.984} & \textbf{0.993} & {0.974} & {0.967} & {0.925} & {0.931} & {0.941} & {0.944} \tabularnewline
{ECGFiveDays} & \textbf{1} & \textbf{1} & \textbf{1} & \textbf{1} & \textbf{1} & {0.984} & \textbf{1} & \textbf{1} & {0.82} \tabularnewline
{FordB} & {0.781} & {0.793} & \underline{0.81} & {0.798} & {0.62} & {0.807} & {0.711} & \textbf{0.823} & {0.662} \tabularnewline
{Ham} & {0.657} & \textbf{0.724} & \underline{0.695} & {0.533} & {0.467} & {0.686} & {0.667} & {0.667} & {0.571} \tabularnewline
{Phoneme} & {0.249} & {0.276} & {0.289} & {0.196} & {0.228} & \underline{0.321} & {0.265} & \textbf{0.382} & {0.305} \tabularnewline
{SwedishLeaf} & {0.925} & {0.914} & \underline{0.931} & {0.925}& {0.792} & {0.928} & {0.922} & \textbf{0.954} & {0.915} \tabularnewline
\bottomrule
\end{tabular}
\end{centering}
\end{table}

\paragraph*{Sparse labeling.}

Taking advantage of their unsupervised training, we show that our representations can be successfully
used on sparsely labeled datasets compared to supervised methods, since only the SVM is restricted to be
learned on the small portion of labeled data. Figure~\ref{fig:SparseLabeling}
shows that an SVM trained on our representations of a randomly chosen labeled set \emph{consistently outperforms
the supervised neural network ResNet} trained on a labeled set of the same size, especially
when the percentage of labeled data is small. For example, with only $1.5\%$ of
labeled data, we achieve an accuracy of $81\%$, against only $26\%$ for ResNet,
equivalent to a random classifier. Moreover, we exceed
$99\%$ of accuracy starting from $11\%$ of labeled data, while ResNet only
achieves this level of accuracy with more than $50\%$ of labeled data.
This shows the relevance of our method in semi-supervised settings, compared to fully supervised methods.

\begin{figure}
	\begin{centering}
	\subfloat[DiatomSizeReduction.]{\includegraphics[width=0.32\columnwidth]{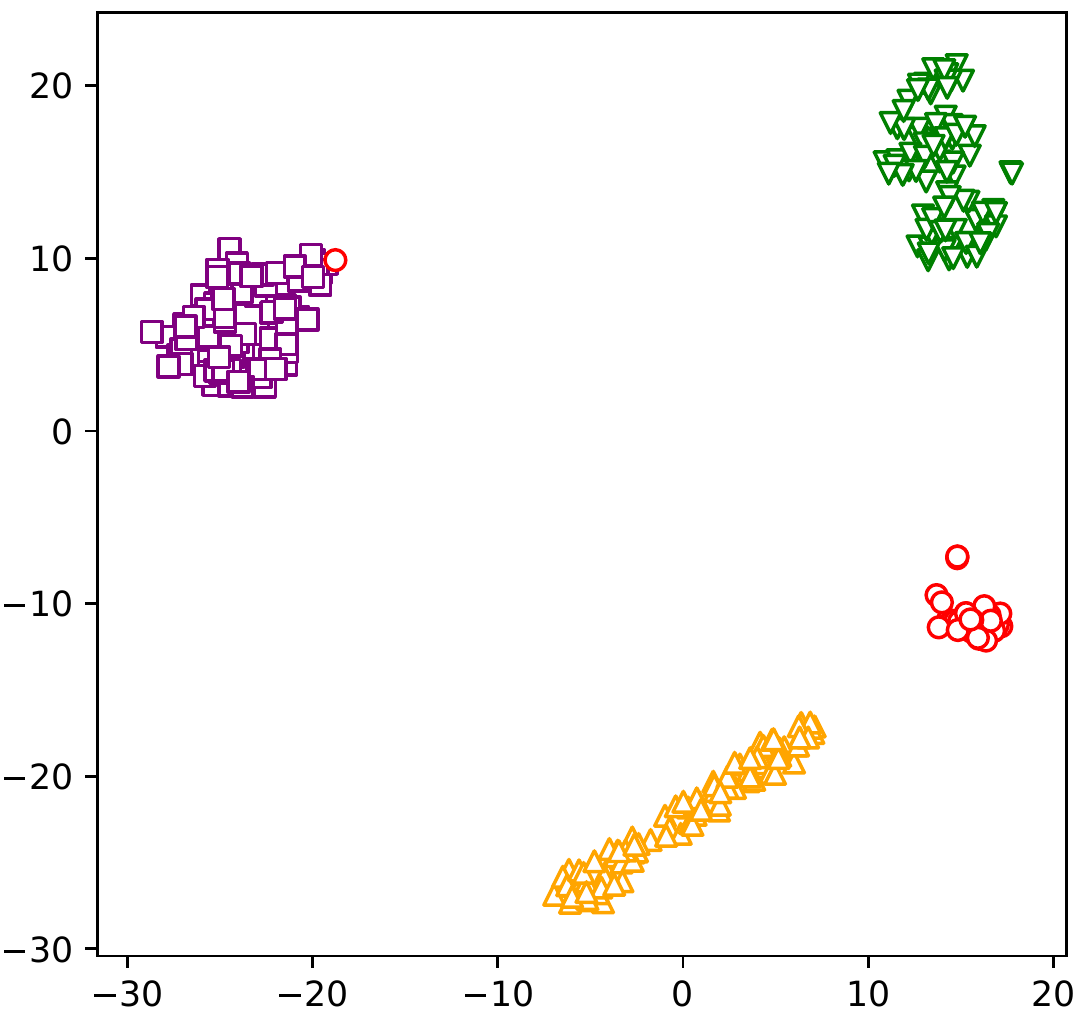}}
	\hfill
	\subfloat[FordB.]{\includegraphics[width=0.32\columnwidth]{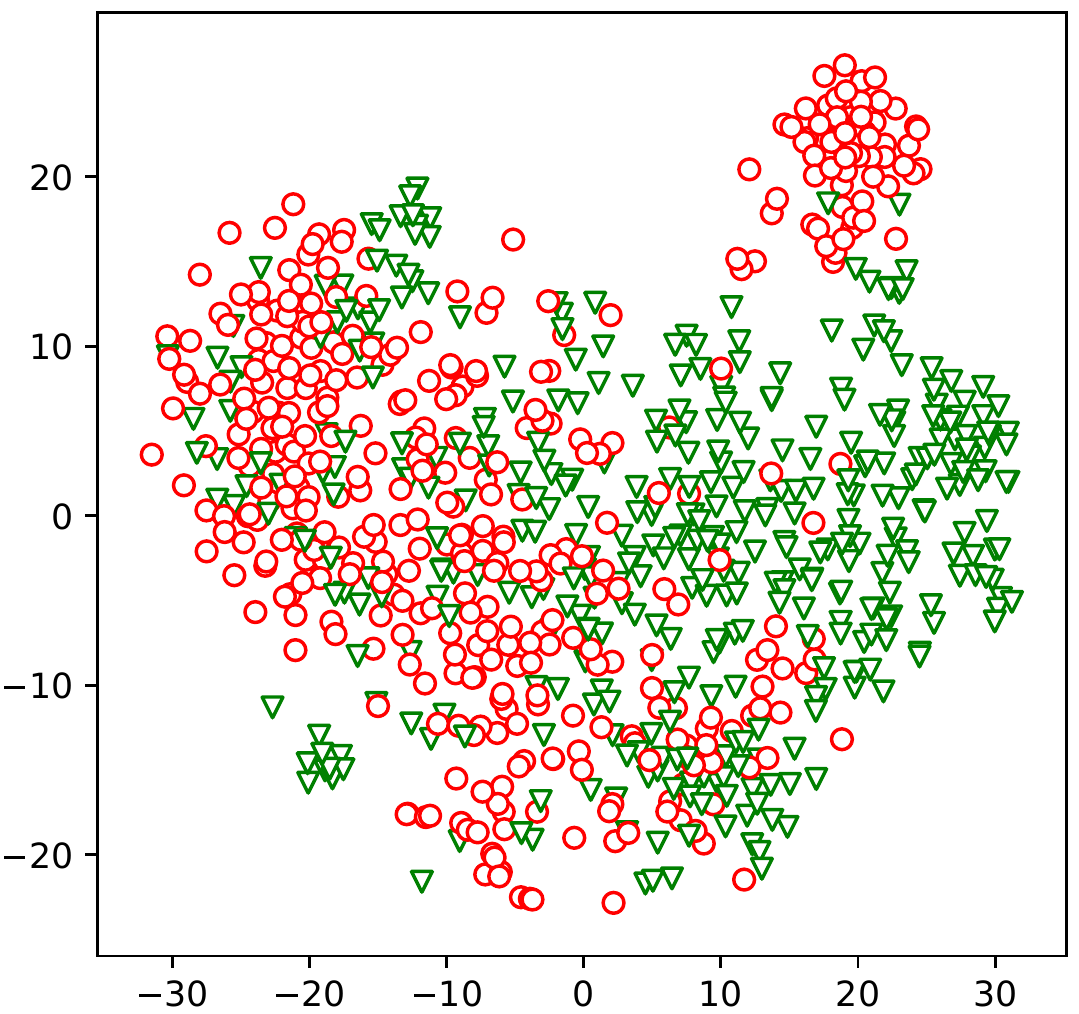}}
	\hfill
	\subfloat[OSULeaf.]{\includegraphics[width=0.32\columnwidth]{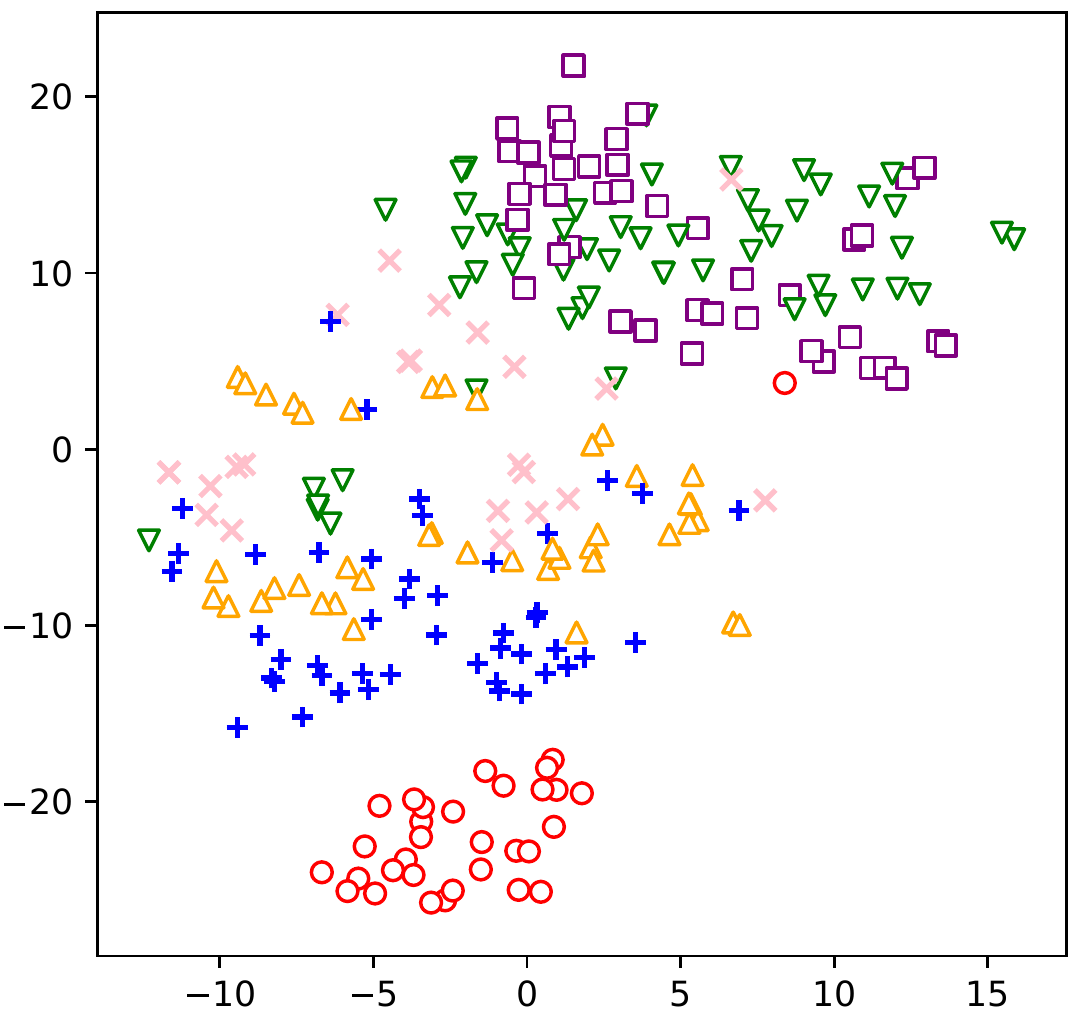}}
	\end{centering}
	\caption{\label{fig:tSNE}Two-dimensional t-SNE \citep{Maaten2008} with perplexity $30$ of the learned representations of three UCR test sets. Elements classes are distinguishable using their respective marker shapes and colors.}
\end{figure}

\paragraph*{Representations metric space.}

Besides being suitable for classification purposes, the learned representations may also be used to
define a meaningful measure between time series. Indeed, we train, instead of an SVM, a one-nearest-neighbor
classifier with respect to the $\ell_2$ distance on the same representations, and compare it
to DTW, which uses the same classifier on the raw time series. As shown in \ref{app:full_univariate_results},
this version of our method outperforms DTW on $66\%$ of the UCR datasets, showing the advantage of
the learned representations even in a non-parametric classification setting. We also include quantitative
experiments to assess the usefulness of comparing time series using the $\ell_2$ distance between their
representations with dimensionality reduction (Figure~\ref{fig:tSNE}) and clustering (Section~\ref{sec:Scalability}
and Figure~\ref{fig:Electricity}) visualizations.

\paragraph*{Transferability.}

We include in the comparisons the classification accuracy for each
dataset of an SVM trained on this dataset using the representations computed by an encoder, which was
trained \emph{on another dataset} (FordA, with $K=5$), to test
the transferability of our representations.
We observe that the scores achieved by this SVM trained on transferred representations are close
to the scores reported when the encoder is trained on the same dataset
as the SVM, showing the \emph{transferability} of our representations
from a dataset to another, and from time series to other time series \emph{with different lengths}.
More generally, this transferability and the
performance of simple classifiers on the representations we learn indicate that they are
\emph{universal} and \emph{easy to make use of}.

\subsubsection{Multivariate Time Series}

To complement our evaluation on the UCR archive which exclusively contains univariate series,
we evaluate our method on multivariate time series. This can be done by simply changing the
number of input filters of the first convolutional layer of the proposed encoder.
We test our method on all 30 datasets of the newly released UEA archive \citep{Bagnall2018}.
Full accuracy scores are presented in the supplementary material, Section~\ref{app:multivariate}, Table~\ref{app_table:multivariate}.

The UEA archive has been designed as a first attempt to provide a standard
archive for multivariate time series classification such as the UCR one for
univariate series. As it has only been released recently, we could not
compare our method to state-of-the-art classifiers for multivariate time series.
However, we provide a comparison with DTW\textsubscript{D} as baseline using results
provided by \citet{Bagnall2018}. DTW\textsubscript{D} (dimension-Dependent DTW)
is a possible extension of DTW in the multivariate setting, and is the best
baseline studied by \citet{Bagnall2018}.
Overall, our method matches or outperforms DTW\textsubscript{D} on $69\%$ of
the UEA datasets, which indicates good performance. As this archive is
destined to grow and evolve in the future, and without further comparisons, no
additional conclusion can be drawn.

\subsection{Evaluation on Long Time Series}
\label{sec:Scalability}

\begin{figure}
\begin{centering}
\includegraphics[width=\textwidth]{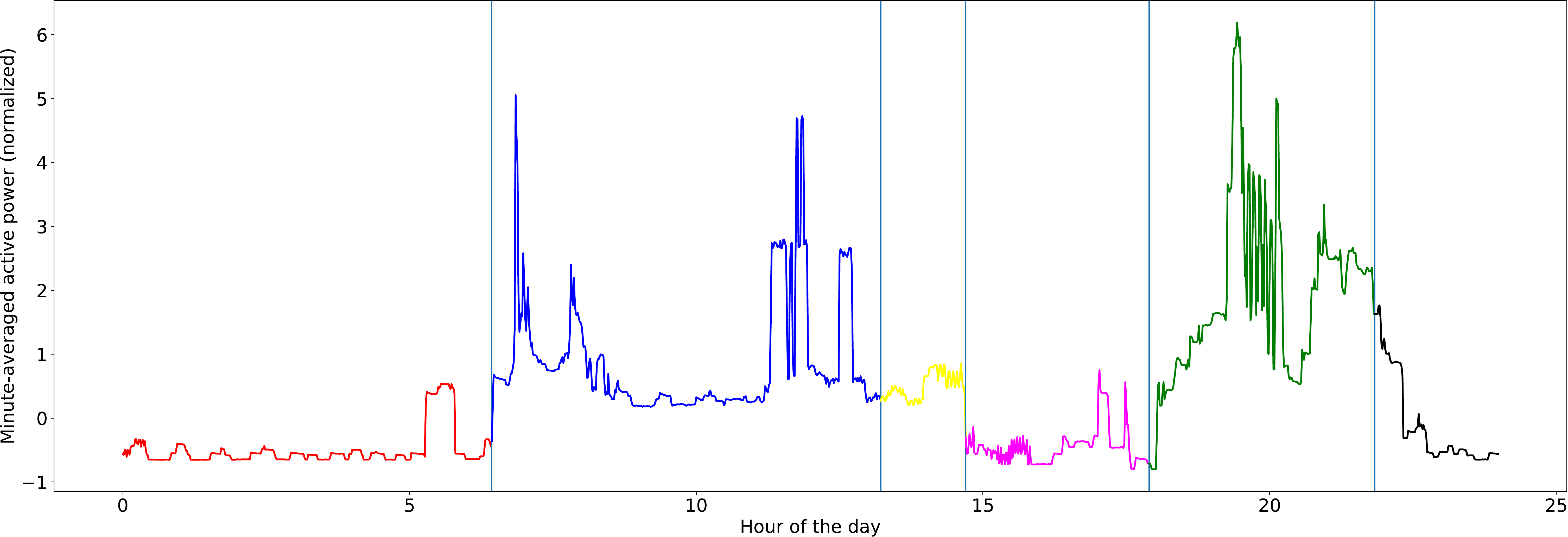}
\caption{\label{fig:Electricity}Minute-averaged electricity consumption for a single day, with respect to the hour of the day. Vertical lines and colors divide the day into six clusters, obtained with $k$-means clustering based on representations computed on a day-long sliding window. The clustering divides the day in meaningful portions (night, morning, afternoon, evening).}
\end{centering}
\end{figure}

We show the \emph{applicability} and \emph{scalability} of our
method on \emph{long} time series without labeling for regression tasks, which
could correspond to an industrial application and complements the performed
tests on the UCR and UEA archives, whose datasets mostly contain short time
series.

The Individual Household Electric Power Consumption (IHEPC) dataset from the UCI
Machine Learning Repository \citep{Dheeru2017} is a single
time series of length $2\,075\,259$ monitoring the minute-averaged
electricity consumption of one French household for four years.
We split this time series into train (first $5\times 10^5$ measurements, approximately a year) and test
(remaining measurements). The encoder is trained over the train time series on a single
Nvidia Tesla P100 GPU in no more than a few hours, showing that our training
procedure is \emph{scalable} to long time series.

We consider the learned encoder on two
regression tasks involving two different input scales. We compute, for each
time step of the time series, the representations of the last window
corresponding to a day ($1\,440$ measurements) and a quarter ($12 \cdot 7 \cdot 1\,440$
measurements) \emph{using the same encoder}. An example of application of the day-long representations is
shown in Figure~\ref{fig:Electricity}.
The considered tasks consist in, for
each time step, predicting the discrepancy between the mean value of the series for
the next period (either a day or quarter) and the one for the previous
period. We compare linear regressors, trained using gradient descent, to minimize
the mean squared error between the prediction and the target, applied
either on the raw time series or the previously computed representations.

\begin{wraptable}{r}{0.5\columnwidth}
\caption{\label{tab:Electricity}Results obtained on the IHEPC dataset.}
\begin{centering}
\scriptsize
\begin{tabular}{cccc}
\toprule
{Task} & {Metric} & {Representations} & {Raw values} \tabularnewline
\midrule
\multirow{2}{*}{{Day}} & {Test MSE} & {$\boldsymbol{8.92\times10^{-2}}$} & {$\boldsymbol{8.92\times10^{-2}}$} \tabularnewline
 & {Wall time} & \textbf{12s} & {3min 1s}\tabularnewline
\midrule
\multirow{2}{*}{{Quarter}} & {Test MSE} & {$7.26\times10^{-2}$} & {$\boldsymbol{6.26\times10^{-2}}$} \tabularnewline
 & {Wall time} & \textbf{9s} & {1h 40min 15s}\tabularnewline
\bottomrule
\end{tabular}
\end{centering}
\end{wraptable}

Results and execution times on an Nvidia Titan Xp GPU are presented in
Table~\ref{tab:Electricity}.\footnote{While acting on representations of the same size,
the quarterly linear regressor is slightly faster than the daily one because
the number of quarters in the considered time series is smaller than the number of
days.} On \emph{both scales of inputs}, our representations
induce only a slightly degraded performance but provide a \emph{large efficiency improvement},
due to their small size compared to the raw time series. This shows that
a single encoder trained to minimize our time-based loss is able to output
representations for different scales of input lengths that are also helpful for other tasks
than classification, corroborating their \emph{universality}.

\section{Conclusion}

We present an unsupervised representation learning method for
time series that is scalable and produces high-quality and easy-to-use
embeddings. They are generated by an encoder formed by dilated convolutions that
admits variable-length inputs, and trained with an efficient triplet loss using
novel time-based negative sampling for time series.
Conducted experiments show that these representations are universal and can easily and efficiently be used
for diverse tasks such as classification, for which we achieve state-of-the-art performance, and
regression.

\newcommand{\acknowledgements}{
\subsubsection*{Acknowledgements}

We would like to acknowledge Patrick Gallinari, Sylvain Lamprier, Mehdi
Lamrayah, Etienne Simon, Valentin Guiguet, Clara Gainon de Forsan de Gabriac,
Eloi Zablocki, Antoine Saporta, Edouard Delasalles, Sidak Pal Singh, Andreas Hug,
Jean-Baptiste Cordonnier, Andreas Loukas and François Fleuret for helpful
comments and discussions.
We thank as well our anonymous reviewers for their constructive suggestions,
\citet{Liljefors2019} for their extensive
and positive reproducibility report on our work, and
all contributors to the datasets and archives we used for this
project \citep{Dau2018, Bagnall2018, Dheeru2017}.
We acknowledge financial support from the SFA-AM ETH Board initiative, the
LOCUST ANR project (ANR-15-CE23-0027) and CLEAR (Center for LEArning
and data Retrieval, joint laboratory with Thales\footnote{\url{https://www.thalesgroup.com}.}).
}

\ifdefined\preprint
\acknowledgements
\fi
\ifdefined\accepted
\acknowledgements
\fi

\bibliography{refs}
\bibliographystyle{icml2019}

\newpage{}

\appendix
\appendixpage
\renewcommand{\thesection}{S\arabic{section}}
\renewcommand{\thetable}{S\arabic{table}}
\renewcommand{\thefigure}{S\arabic{figure}}
\renewcommand{\thefootnote}{S\arabic{footnote}}
\setcounter{figure}{0}
\setcounter{table}{0}
\setcounter{footnote}{0}

In these appendices, we provide our detailed training procedure for
classification tasks, choices of hyperparameters, as well as the full
experimental results of our method, compared to those of concurrent methods.
Section~\ref{app:detailed_training} explains and discusses the exact training
process for classification tasks.
Section~\ref{app:hyper_params} details the choices of hyperparameters in all
presented experiments. Section~\ref{app:full_univariate_results} reports
accuracy scores of all variants of our method on the whole UCR archive \citep{Dau2018}, as well
as comparisons with concurrent methods, when available.
Section~\ref{app:multivariate} provides accuracy scores for our method on the
whole UEA archive \citep{Bagnall2018}.
Finally, Section~\ref{app:EncoderDiscussion} discusses the importance of the choice of
encoder by providing results obtained with our training procedure using an LSTM as encoder.

\section{Training Details}
\label{app:detailed_training}

\subsection{Input Preprocessing}

We preprocess datasets of the UCR archive that were not already normalized, as well as the IHEPC dataset,
so that the set of time series values for each dataset has zero mean and unit
variance.
For each UEA dataset, each dimension of the time series was preprocessed
independently from the other dimensions by normalizing in the same way its mean
and variance.

\subsection{SVM Training}
\label{app:SVM}

In order to train an SVM on the computed representations of the elements of the
train set, we perform a hyperparameter optimization for the penalty $C$ of
the error term of the SVM by cross-validating it over the
representations of the train set, thus only using the train labels.
Note that if the train set or the number of training
samples per class are too small, we
choose a penalty $C=\infty$ for the SVM (which corresponds to no regularization).

\subsection{Behavior of the Learned Representations through Training}

\begin{figure}
\begin{centering}
\includegraphics[width=0.8\textwidth]{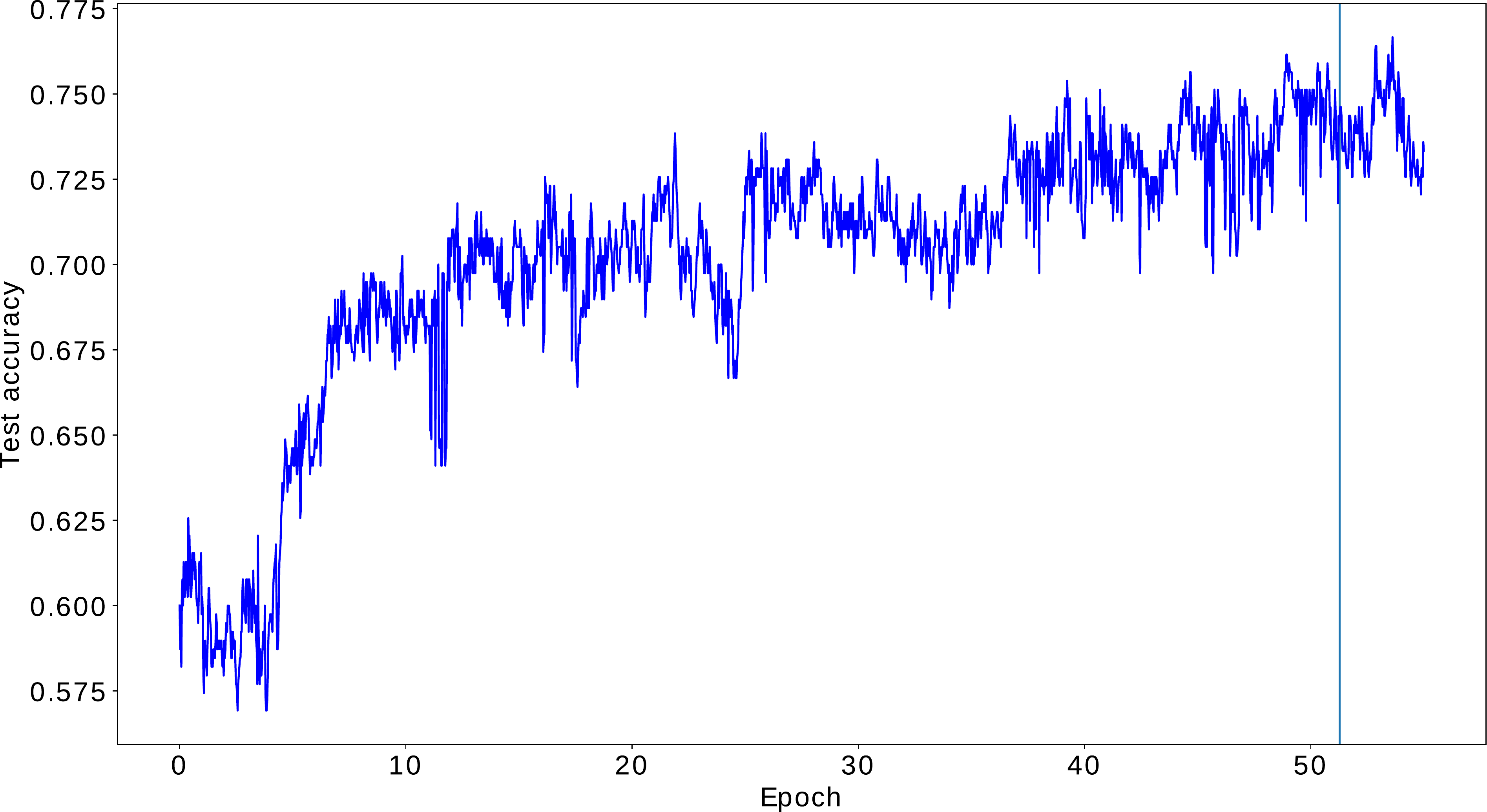}
\caption{\label{fig:LearningCurve}Evolution of the test accuracy during the training of the encoder on the CricketX dataset from the UCR archive (with $K=10$), with respect to the number of completed epochs. The test labels were only used for monitoring purposes and the test accuracy was computed after each mini-batch optimization. The vertical line marks the epoch at which $2000$ optimization steps were performed, at which point training is stopped in our tests. Test accuracy clearly increases during training.}
\end{centering}
\end{figure}

\paragraph*{Classification accuracy evolution during training.}
As shown in Figure~\ref{fig:LearningCurve}, our unsupervised training clearly
makes the classification accuracy of the trained SVM increase with the number of
optimization steps.

\paragraph*{Numerical stability.}
The \emph{Risk} $R$  is defined as the expectation (taken over the random selection of the sequences $ \parentheses*{\xx^{\mathrm{ref}}, \xx^{\mathrm{pos}}, \xx^{\mathrm{neg}}} $) of the loss defined in Equation~\eqref{eq:def_loss}.  This risk may decrease if all the representations $\ff\parentheses*{\cdot ,\ttheta} $ are scaled by a positive large number.  For example, if for some $ \ttheta_0 $, for (almost surely) any sequences $ \parentheses*{\xx^{\mathrm{ref}}, \xx^{\mathrm{pos}}, \xx^{\mathrm{neg}}} $, $ \ff\parentheses*{\xx^{\mathrm{ref}},\ttheta_0}^\top \ff\parentheses*{\xx^{\mathrm{pos}},\ttheta_0} \geq 0$ and $
\ff\parentheses*{\xx^{\mathrm{ref}},\ttheta_0}^\top \ff\parentheses*{\xx^{\mathrm{neg}},\ttheta_0}\le 0$, then
\begin{multline}
R\parentheses*{\lambda, \ttheta_0} := \mathbb{E}_{\xx^{\mathrm{ref}}, \xx^{\mathrm{pos}}, \xx^{\mathrm{neg}}} \lbracket*{-\log\parentheses*{\sigma\parentheses*{\lambda^{2} \ff\parentheses*{\xx^{\mathrm{ref}},\ttheta_0}^\top \ff\parentheses*{\xx^{\mathrm{pos}},\ttheta_0}}}}\\
\rbracket*{-\log\parentheses*{\sigma\parentheses*{- \lambda^2 \ff\parentheses*{\xx^{\mathrm{ref}},\ttheta_0}^\top \ff\parentheses*{\xx^{\mathrm{neg}},\ttheta_0}}}}
\end{multline}
is a decreasing function of $ \lambda $, thus $ \lambda $ could diverge to infinity in order to minimize the loss. In other words, the parameters in $ \ttheta_0 $   corresponding to the last linear layer could be linearly scaled up, and representations would ``explode'' (their norm would always increase through training). Such a phenomenon is not observed in practice, as the mean representation Euclidean norm lies around $20$. There are two possible explanations for that: either the condition above is not satisfied  (more generally, the loss is not reduced by increasing the representations) or the use of the sigmoid function, that has vanishing gradients, results in an increase of the representations that is too slow to be observed, or negligible with respect to other weight updates during optimization.

\section{Hyperparameters}
\label{app:hyper_params}

\subsection{Influence of $K$}

As mentioned in Section~\ref{sec:ExperimentalResults}, $K$ can have a
significant impact on the performance of the encoder. We notably observed that
$K=1$ leads to statistically significantly lower scores compared to scores
obtained when trained with $K > 1$ on the UCR datasets, justifying the use of
several negative examples during training. We did not observe any
clear statistical difference between other values of $K$ on the whole archive;
however, we noticed important differences between different values of $K$ when
studying individual datasets. Therefore, we chose to combine several
encoders trained with different values of $K$ in order to avoid selecting it as a
fixed hyperparameter.

\subsection{Detailed Choices of Hyperparameters}

We train our models with the following parameters for time series classification.
Note that \emph{no hyperparameter optimization} was performed on the encoder hyperparameters.

\begin{itemize}
\item Optimizer: Adam \citep{Kingma2015} with learning rate $\alpha=0.001$
and decay rates $\beta=\parentheses*{0.9,0.999}$.
\item SVM: penalty $C\in\braces*{10^i \mid i \in \lrbrackets*{-4, 4}} \cup \braces*{\infty} $.
\item Encoder training:
\begin{itemize}
\item number of negative samples: $K\in\braces*{1,2,5,10}$ for univariate time series, $K\in\braces*{5,10, 20}$ for multivariate ones;
\item batch size: $10$;
\item number of optimizations steps: $2000$ for $K \geq 10$ (i.e., $20$ epochs for a dataset of size $1000$), $1500$ otherwise.
\end{itemize}
\item Architecture:
\begin{itemize}
\item number of channels in the intermediary layers of the causal network:
$40$;
\item number of layers (depth of the causal network): $10$;
\item kernel size of all convolutions: $3$;
\item negative slope of the leaky ReLU activation: $0.01$;
\item number of output channels of the causal network (before max pooling): $320$;
\item dimension of the representations: $160$.
\end{itemize}
\end{itemize}
For the Individual Household Electric Power Consumption dataset, changes are the following:
\begin{itemize}
\item number of negative samples: $K=10$;
\item batch size: $1$;
\item number of optimization steps: $400$;
\item number of channels in the intermediary layers of the causal network:
$30$;
\item number of output channels of the causal network (before max pooling): $160$;
\item dimension of the representations: $80$.
\end{itemize}

\section{Univariate Time Series}
\label{app:full_univariate_results}

\begin{figure}
\begin{centering}
\includegraphics[width=0.8\columnwidth]{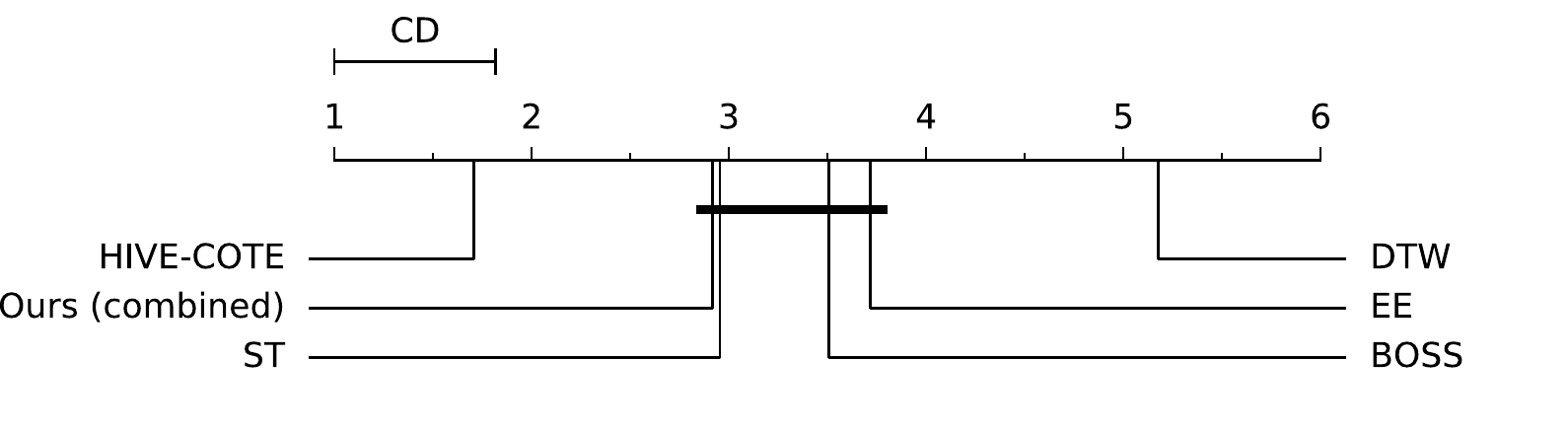}
\par\end{centering}
\caption{\label{fig:CriticalDifferenceDiagram}Critical difference diagram
of the average ranks of the compared classifiers for the Nemenyi test,
obtained with Orange \citep{Demsar2013}.}
\end{figure}

\begin{figure}
\begin{centering}
\includegraphics[width=0.65\columnwidth]{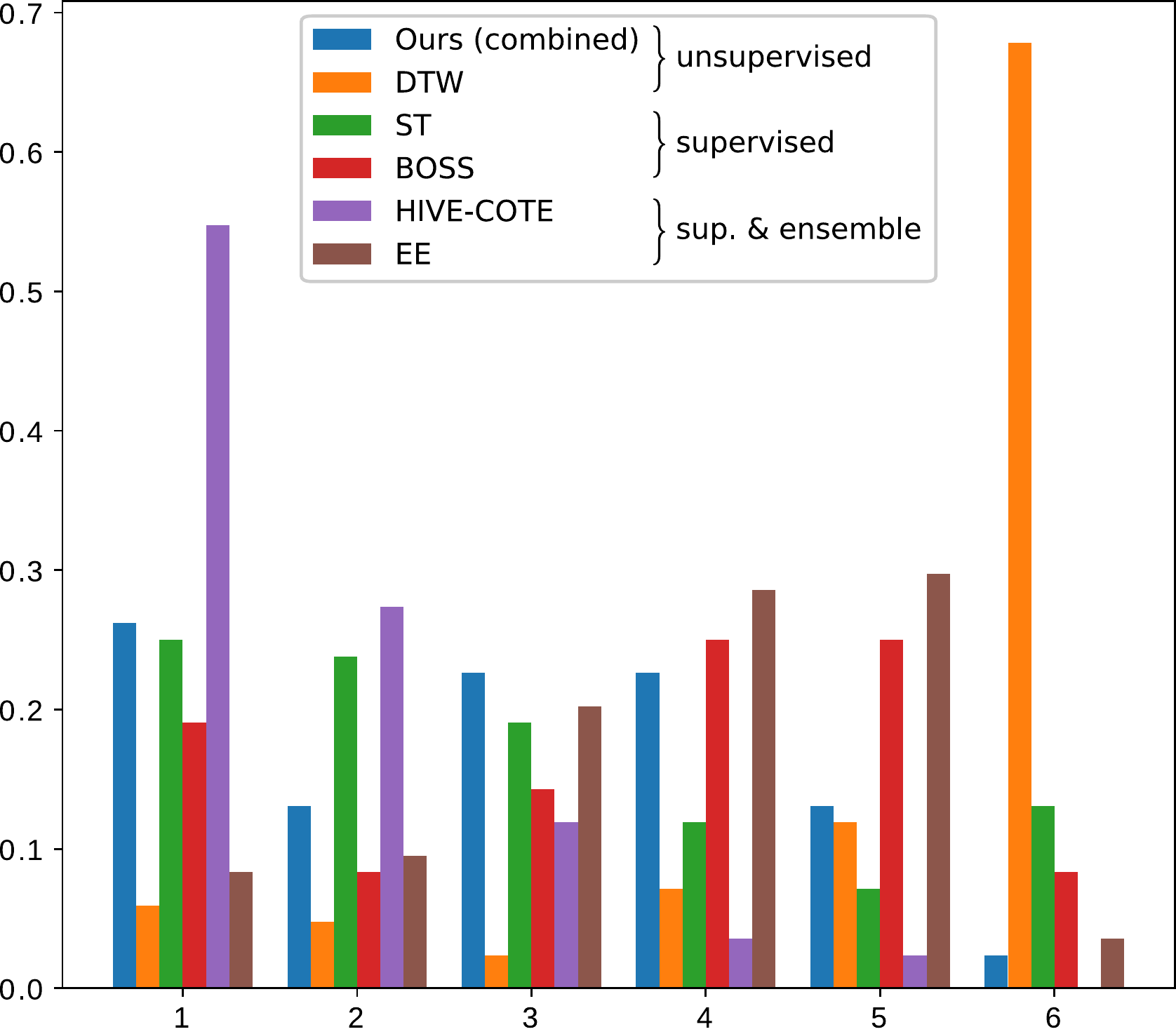}
\par\end{centering}
\caption{\label{fig:Histogram}Distribution of ranks of compared methods for the
first 85 UCR datasets.}
\end{figure}

Full results corresponding to the first 85 UCR datasets for our method are presented in Table~\ref{app_table:our_results_ucr}, while comparisons with DTW, ST, BOSS, HIVE-COTE and EE are shown in Figures~\ref{fig:CriticalDifferenceDiagram} and~\ref{fig:Histogram} and Table~\ref{app_table:compar_results_ucr},\footnote{Scores taken from \url{http://www.timeseriesclassification.com/singleTrainTest.csv}.} and comparisons with
ResNet,\footnote{Scores taken from \url{https://github.com/hfawaz/dl-4-tsc/blob/master/results/results-uea.csv} (first iteration).}
TimeNet and RWS are shown in Table~\ref{app_table:comparison_to_others_deep}. Table~\ref{app_table:our_results_ucr_newest} compiles the results of our method and of DTW\footnote{Scores taken from \url{https://www.cs.ucr.edu/~eamonn/time_series_data_2018/}.} for the newest 43 UCR datasets (except DodgerLoopDay, DodgerLoopGame and DodgerLoopWeekend which contain missing values).

\paragraph*{Standard deviation.}
All UCR datasets are provided with a unique train / test split that we used in
our experiments. Compared techniques (DTW, ST, BOSS, HIVE-COTE and EE) were
also tested on 100 random train / test splits of these datasets by
\citet{Bagnall2017} to produce a strong state-of-the-art evaluation, but we
did not perform similar resamples as this is beyond the scope of this work and
would require much more computations. Note that the scores for these methods
used in this article are the ones corresponding to the original train / test
split of the datasets.

As our method is based on random sampling, the reported scores may vary
depending on the random seed. While we do not report standard deviation, the
large number of tested datasets prevents large statistical error in the global
evaluation of our method. The order of magnitude of accuracy variation between
different runs of the combined version of our method is below $0.01$ (for instance,
on four different runs, the corresponding standard variations for, respectively,
datasets DiatomSizeReduction, CricketX and UWaveGestureLibraryX are $0.0056$, $0.0091$ and $0.0053$).

\newpage{}

\begingroup
\scriptsize
\begin{longtable}[c]{lccccccc}
\caption{\label{app_table:our_results_ucr}Accuracy scores of variants of our method on the first 85 UCR datasets. ``Combined ($1$-NN)'' corresponds to learning a one-nearest-neighbor classifier, instead of an SVM, on the combined representations. Bold scores indicate the best performing method.}
\tabularnewline
\toprule
\multirow{3}{*}[-\dimexpr \aboverulesep + \belowrulesep + \cmidrulewidth]{{Dataset}} & \multicolumn{7}{c}{{Unsupervised}} \tabularnewline
\cmidrule(lr){2-8}
& \multicolumn{7}{c}{{Ours}} \tabularnewline
\cmidrule(lr){2-8}
& {$K=1$} & {$K=2$} & {$K=5$} & {$K=10$} & {Combined} & {Combined ($1$-NN)} & {FordA} \tabularnewline
\midrule
\endhead
{Adiac} & {0.734} & {0.711} & {0.703} & {0.675} & {0.716} & {0.645} & \textbf{0.76} \tabularnewline
{ArrowHead} & \textbf{0.869} & {0.829} & {0.754} & {0.766} & {0.829} & {0.817} & {0.817} \tabularnewline
{Beef} & \textbf{0.733} & {0.567} & {0.7} & {0.667} & {0.7} & {0.6} & {0.667} \tabularnewline
{BeetleFly} & \textbf{0.9} & {0.8} & \textbf{0.9} & {0.8} & \textbf{0.9} & {0.8} & {0.8} \tabularnewline
{BirdChicken} & {0.7} & {0.8} & \textbf{0.9} & {0.85} & {0.8} & {0.75} & \textbf{0.9} \tabularnewline
{Car} & {0.75} & {0.767} & {0.633} & {0.833} & {0.817} & {0.8} & \textbf{0.85} \tabularnewline
{CBF} & {0.982} & {0.991} & {0.99} & {0.983} & \textbf{0.994} & {0.978} & {0.988} \tabularnewline
{ChlorineConcentration} & {0.719} & {0.747} & {0.739} & {0.749} & \textbf{0.782} & {0.588} & {0.688} \tabularnewline
{CinCECGTorso} & {0.702} & \textbf{0.747} & {0.682} & {0.713} & {0.74} & {0.693} & {0.638} \tabularnewline
{Coffee} & {0.964} & \textbf{1} & \textbf{1} & \textbf{1} & \textbf{1} & \textbf{1} & \textbf{1} \tabularnewline
{Computers} & \textbf{0.688} & {0.644} & {0.676} & {0.664} & {0.628} & {0.604} & {0.648} \tabularnewline
{CricketX} & {0.736} & {0.71} & {0.7} & {0.713} & \textbf{0.777} & {0.741} & {0.682} \tabularnewline
{CricketY} & {0.682} & {0.664} & {0.695} & {0.728} & \textbf{0.767} & {0.664} & {0.667} \tabularnewline
{CricketZ} & {0.721} & {0.71} & {0.726} & {0.708} & \textbf{0.764} & {0.723} & {0.656} \tabularnewline
{DiatomSizeReduction} & {0.99} & {0.987} & \textbf{0.993} & {0.984} & \textbf{0.993} & {0.967} & {0.974} \tabularnewline
{DistalPhalanxOutlineCorrect} & {0.761} & {0.746} & \textbf{0.775} & \textbf{0.775} & {0.768} & {0.757} & {0.764} \tabularnewline
{DistalPhalanxOutlineAgeGroup} & {0.719} & \textbf{0.748} & {0.719} & {0.727} & {0.734} & {0.683} & {0.727} \tabularnewline
{DistalPhalanxTW} & \textbf{0.698} & {0.676} & {0.662} & {0.676} & {0.676} & {0.669} & {0.669} \tabularnewline
{Earthquakes} & \textbf{0.748} & \textbf{0.748} & \textbf{0.748} & \textbf{0.748} & \textbf{0.748} & {0.64} & \textbf{0.748} \tabularnewline
{ECG200} & {0.87} & {0.9} & {0.86} & \textbf{0.94} & {0.9} & {0.85} & {0.83} \tabularnewline
{ECG5000} & {0.939} & {0.939} & {0.937} & {0.933} & {0.936} & {0.925} & \textbf{0.94} \tabularnewline
{ECGFiveDays} & \textbf{1} & \textbf{1} & \textbf{1} & \textbf{1} & \textbf{1} & {0.999} & \textbf{1} \tabularnewline
{ElectricDevices} & {0.709} & {0.7} & {0.712} & {0.707} & \textbf{0.732} & {0.646} & {0.676} \tabularnewline
{FaceAll} & {0.764} & \textbf{0.81} & {0.733} & {0.786} & {0.802} & {0.75} & {0.734} \tabularnewline
{FaceFour} & {0.807} & {0.864} & {0.795} & \textbf{0.92} & {0.875} & {0.864} & {0.83} \tabularnewline
{FacesUCR} & {0.885} & {0.871} & {0.886} & {0.884} & \textbf{0.918} & {0.86} & {0.835} \tabularnewline
{FiftyWords} & {0.763} & {0.734} & {0.727} & {0.732} & \textbf{0.78} & {0.716} & {0.745} \tabularnewline
{Fish} & {0.903} & {0.909} & {0.891} & {0.891} & {0.88} & {0.823} & \textbf{0.96} \tabularnewline
{FordA} & {0.923} & {0.922} & {0.927} & {0.928} & \textbf{0.935} & {0.863} & {0.927} \tabularnewline
{FordB} & {0.786} & {0.788} & {0.781} & {0.793} & \textbf{0.81} & {0.748} & {0.798} \tabularnewline
{GunPoint} & {0.953} & {0.987} & {0.987} & {0.98} & \textbf{0.993} & {0.833} & {0.987} \tabularnewline
{Ham} & {0.648} & {0.686} & {0.657} & \textbf{0.724} & {0.695} & {0.533} & {0.533} \tabularnewline
{HandOutlines} & \textbf{0.922} & {0.919} & {0.908} & \textbf{0.922} & \textbf{0.922} & {0.832} & {0.919} \tabularnewline
{Haptics} & {0.445} & {0.435} & {0.432} & \textbf{0.49} & {0.455} & {0.354} & {0.474} \tabularnewline
{Herring} & \textbf{0.609} & {0.594} & {0.578} & {0.594} & {0.578} & {0.563} & {0.578} \tabularnewline
{InlineSkate} & {0.425} & {0.429} & {0.427} & {0.371} & \textbf{0.447} & {0.4} & {0.444} \tabularnewline
{InsectWingbeatSound} & {0.61} & {0.592} & {0.617} & {0.597} & \textbf{0.623} & {0.506} & {0.599} \tabularnewline
{ItalyPowerDemand} & {0.94} & {0.927} & {0.928} & \textbf{0.954} & {0.925} & {0.942} & {0.929} \tabularnewline
{LargeKitchenAppliances} & {0.797} & {0.827} & {0.843} & {0.789} & \textbf{0.848} & {0.757} & {0.765} \tabularnewline
{Lightning2} & {0.869} & {0.836} & {0.852} & {0.869} & \textbf{0.918} & {0.885} & {0.787} \tabularnewline
{Lightning7} & {0.795} & \textbf{0.822} & \textbf{0.822} & {0.795} & {0.795} & {0.795} & {0.74} \tabularnewline
{Mallat} & {0.962} & {0.931} & {0.947} & {0.951} & \textbf{0.964} & {0.944} & {0.916} \tabularnewline
{Meat} & {0.917} & {0.867} & {0.867} & \textbf{0.95} & \textbf{0.95} & {0.9} & {0.867} \tabularnewline
{MedicalImages} & {0.738} & {0.768} & {0.77} & {0.75} & \textbf{0.784} & {0.693} & {0.725} \tabularnewline
{MiddlePhalanxOutlineCorrect} & {0.749} & {0.818} & {0.777} & \textbf{0.825} & {0.814} & {0.722} & {0.787} \tabularnewline
{MiddlePhalanxOutlineAgeGroup} & {0.617} & \textbf{0.662} & {0.656} & {0.656} & {0.656} & {0.506} & {0.623} \tabularnewline
{MiddlePhalanxTW} & {0.604} & \textbf{0.61} & \textbf{0.61} & {0.591} & \textbf{0.61} & {0.513} & {0.584} \tabularnewline
{MoteStrain} & \textbf{0.875} & {0.854} & {0.867} & {0.851} & {0.871} & {0.853} & {0.823} \tabularnewline
{NonInvasiveFetalECGThorax1} & {0.912} & {0.911} & {0.904} & {0.878} & {0.91} & {0.798} & \textbf{0.925} \tabularnewline
{NonInvasiveFetalECGThorax2} & {0.925} & {0.925} & {0.918} & {0.919} & {0.927} & {0.82} & \textbf{0.93} \tabularnewline
{OliveOil} & {0.867} & {0.833} & {0.867} & {0.867} & \textbf{0.9} & {0.833} & \textbf{0.9} \tabularnewline
{OSULeaf} & {0.719} & {0.694} & {0.793} & {0.76} & \textbf{0.831} & {0.636} & {0.736} \tabularnewline
{PhalangesOutlinesCorrect} & \textbf{0.807} & {0.796} & {0.795} & {0.784} & {0.801} & {0.752} & {0.784} \tabularnewline
{Phoneme} & {0.264} & {0.265} & {0.249} & {0.276} & \textbf{0.289} & {0.197} & {0.196} \tabularnewline
{Plane} & {0.99} & \textbf{1} & {0.99} & {0.99} & {0.99} & \textbf{1} & {0.981} \tabularnewline
{ProximalPhalanxOutlineCorrect} & \textbf{0.869} & {0.863} & {0.856} & {0.859} & {0.859} & {0.801} & \textbf{0.869} \tabularnewline
{ProximalPhalanxOutlineAgeGroup} & {0.849} & \textbf{0.859} & {0.844} & {0.844} & {0.854} & {0.805} & {0.839} \tabularnewline
{ProximalPhalanxTW} & \textbf{0.824} & {0.815} & {0.761} & {0.771} & \textbf{0.824} & {0.717} & {0.785} \tabularnewline
{RefrigerationDevices} & {0.531} & {0.507} & {0.547} & {0.515} & {0.517} & {0.475} & \textbf{0.555} \tabularnewline
{ScreenType} & {0.408} & {0.411} & \textbf{0.427} & {0.416} & {0.413} & {0.389} & {0.384} \tabularnewline
{ShapeletSim} & \textbf{0.894} & {0.5} & {0.628} & {0.672} & {0.817} & {0.772} & {0.517} \tabularnewline
{ShapesAll} & {0.847} & {0.84} & {0.857} & {0.848} & \textbf{0.875} & {0.823} & {0.837} \tabularnewline
{SmallKitchenAppliances} & {0.68} & {0.667} & {0.715} & {0.677} & {0.715} & {0.619} & \textbf{0.731} \tabularnewline
{SonyAIBORobotSurface1} & \textbf{0.93} & {0.89} & {0.85} & {0.902} & {0.897} & {0.825} & {0.84} \tabularnewline
{SonyAIBORobotSurface2} & {0.885} & {0.933} & {0.928} & {0.889} & \textbf{0.934} & {0.885} & {0.832} \tabularnewline
{StarLightCurves} & {0.96} & {0.966} & {0.958} & {0.964} & {0.965} & {0.893} & \textbf{0.968} \tabularnewline
{Strawberry} & {0.951} & {0.946} & \textbf{0.954} & \textbf{0.954} & {0.946} & {0.903} & {0.946} \tabularnewline
{SwedishLeaf} & {0.907} & {0.925} & {0.925} & {0.914} & \textbf{0.931} & {0.891} & {0.925} \tabularnewline
{Symbols} & {0.937} & {0.931} & \textbf{0.965} & {0.963} & \textbf{0.965} & {0.933} & {0.945} \tabularnewline
{SyntheticControl} & {0.98} & {0.983} & \textbf{0.987} & \textbf{0.987} & {0.983} & {0.977} & {0.977} \tabularnewline
{ToeSegmentation1} & {0.868} & \textbf{0.961} & {0.93} & {0.939} & {0.952} & {0.851} & {0.899} \tabularnewline
{ToeSegmentation2} & {0.869} & {0.892} & {0.838} & \textbf{0.9} & {0.885} & \textbf{0.9} & \textbf{0.9} \tabularnewline
{Trace} & \textbf{1} & \textbf{1} & \textbf{1} & {0.99} & \textbf{1} & \textbf{1} & \textbf{1} \tabularnewline
{TwoLeadECG} & {0.996} & {0.991} & {0.996} & \textbf{0.999} & {0.997} & {0.988} & {0.993} \tabularnewline
{TwoPatterns} & {0.998} & \textbf{1} & \textbf{1} & {0.999} & \textbf{1} & {0.998} & {0.992} \tabularnewline
{UWaveGestureLibraryX} & {0.795} & {0.791} & {0.806} & {0.785} & \textbf{0.811} & {0.762} & {0.784} \tabularnewline
{UWaveGestureLibraryY} & {0.716} & {0.717} & {0.702} & {0.71} & \textbf{0.735} & {0.666} & {0.697} \tabularnewline
{UWaveGestureLibraryZ} & {0.738} & {0.735} & {0.741} & {0.757} & \textbf{0.759} & {0.679} & {0.729} \tabularnewline
{UWaveGestureLibraryAll} & {0.893} & {0.887} & {0.903} & {0.896} & \textbf{0.941} & {0.838} & {0.865} \tabularnewline
{Wafer} & {0.991} & \textbf{0.995} & {0.993} & {0.992} & {0.993} & {0.987} & \textbf{0.995} \tabularnewline
{Wine} & {0.704} & {0.815} & {0.852} & {0.815} & \textbf{0.87} & {0.5} & {0.685} \tabularnewline
{WordSynonyms} & {0.63} & {0.646} & {0.676} & {0.691} & \textbf{0.704} & {0.633} & {0.641} \tabularnewline
{Worms} & {0.662} & \textbf{0.74} & {0.688} & {0.727} & {0.714} & {0.597} & {0.688} \tabularnewline
{WormsTwoClass} & {0.753} & {0.766} & {0.74} & {0.792} & \textbf{0.818} & {0.805} & {0.753} \tabularnewline
{Yoga} & {0.824} & {0.854} & {0.831} & {0.837} & \textbf{0.878} & {0.837} & {0.828} \tabularnewline
\bottomrule
\end{longtable}
\endgroup

\newpage

\begingroup
\scriptsize
\begin{longtable}[c]{lcccccc}
\caption{\label{app_table:compar_results_ucr}Accuracy scores of the combined version of our method compared with
those of DTW (unsupervised), ST and BOSS (supervised) and HIVE-COTE and EE (supervised ensemble methods), on the first 85 UCR datasets
(results on the full archive were not available for comparisons).
Bold scores indicate the best performing method.}
\tabularnewline
\toprule
\multirow{3}{*}[-\dimexpr \aboverulesep + \belowrulesep + \cmidrulewidth]{{Dataset}} & \multicolumn{2}{c}{{Unsupervised}} & \multicolumn{4}{c}{{Supervised}} \tabularnewline
\cmidrule(lr){2-3} \cmidrule(lr){4-7}
& {Ours} & \multirow{2}{*}[-0.5\dimexpr \aboverulesep + \belowrulesep + \cmidrulewidth]{{DTW}} & \multirow{2}{*}[-0.5\dimexpr \aboverulesep + \belowrulesep + \cmidrulewidth]{{ST}} & \multirow{2}{*}[-0.5\dimexpr \aboverulesep + \belowrulesep + \cmidrulewidth]{{BOSS}} & \multicolumn{2}{c}{{Ensemble}} \tabularnewline
\cmidrule(lr){2-2} \cmidrule(lr){6-7}
& {Combined} & & & & {HIVE-COTE} & {EE} \tabularnewline
\midrule
\endhead
{Adiac} & {0.716} & {0.604} & {0.783} & {0.765} & \textbf{0.811} & {0.665} \tabularnewline
{ArrowHead} & {0.829} & {0.703} & {0.737} & {0.834} & \textbf{0.863} & {0.811} \tabularnewline
{Beef} & {0.7} & {0.633} & {0.9} & {0.8} & \textbf{0.933} & {0.633} \tabularnewline
{BeetleFly} & {0.9} & {0.7} & {0.9} & {0.9} & \textbf{0.95} & {0.75} \tabularnewline
{BirdChicken} & {0.8} & {0.75} & {0.8} & \textbf{0.95} & {0.85} & {0.8} \tabularnewline
{Car} & {0.817} & {0.733} & \textbf{0.917} & {0.833} & {0.867} & {0.833} \tabularnewline
{CBF} & {0.994} & {0.997} & {0.974} & {0.998} & \textbf{0.999} & {0.998} \tabularnewline
{ChlorineConcentration} & \textbf{0.782} & {0.648} & {0.7} & {0.661} & {0.712} & {0.656} \tabularnewline
{CinCECGTorso} & {0.74} & {0.651} & {0.954} & {0.887} & \textbf{0.996} & {0.942} \tabularnewline
{Coffee} & \textbf{1} & \textbf{1} & {0.964} & \textbf{1} & \textbf{1} & \textbf{1} \tabularnewline
{Computers} & {0.628} & {0.7} & {0.736} & {0.756} & \textbf{0.76} & {0.708} \tabularnewline
{CricketX} & {0.777} & {0.754} & {0.772} & {0.736} & \textbf{0.823} & {0.813} \tabularnewline
{CricketY} & {0.767} & {0.744} & {0.779} & {0.754} & \textbf{0.849} & {0.805} \tabularnewline
{CricketZ} & {0.764} & {0.754} & {0.787} & {0.746} & \textbf{0.831} & {0.782} \tabularnewline
{DiatomSizeReduction} & \textbf{0.993} & {0.967} & {0.925} & {0.931} & {0.941} & {0.944} \tabularnewline
{DistalPhalanxOutlineCorrect} & {0.768} & {0.717} & \textbf{0.775} & {0.728} & {0.772} & {0.728} \tabularnewline
{DistalPhalanxOutlineAgeGroup} & {0.734} & \textbf{0.77} & \textbf{0.77} & {0.748} & {0.763} & {0.691} \tabularnewline
{DistalPhalanxTW} & {0.676} & {0.59} & {0.662} & {0.676} & \textbf{0.683} & {0.647} \tabularnewline
{Earthquakes} & \textbf{0.748} & {0.719} & {0.741} & \textbf{0.748} & \textbf{0.748} & {0.741} \tabularnewline
{ECG200} & \textbf{0.9} & {0.77} & {0.83} & {0.87} & {0.85} & {0.88} \tabularnewline
{ECG5000} & {0.936} & {0.924} & {0.944} & {0.941} & \textbf{0.946} & {0.939} \tabularnewline
{ECGFiveDays} & \textbf{1} & {0.768} & {0.984} & \textbf{1} & \textbf{1} & {0.82} \tabularnewline
{ElectricDevices} & {0.732} & {0.602} & {0.747} & \textbf{0.799} & {0.77} & {0.663} \tabularnewline
{FaceAll} & {0.802} & {0.808} & {0.779} & {0.782} & {0.803} & \textbf{0.849} \tabularnewline
{FaceFour} & {0.875} & {0.83} & {0.852} & \textbf{1} & {0.955} & {0.909} \tabularnewline
{FacesUCR} & {0.918} & {0.905} & {0.906} & {0.957} & \textbf{0.963} & {0.945} \tabularnewline
{FiftyWords} & {0.78} & {0.69} & {0.705} & {0.705} & {0.809} & \textbf{0.82} \tabularnewline
{Fish} & {0.88} & {0.823} & \textbf{0.989} & \textbf{0.989} & \textbf{0.989} & {0.966} \tabularnewline
{FordA} & {0.935} & {0.555} & {0.971} & {0.93} & \textbf{0.964} & {0.738} \tabularnewline
{FordB} & {0.81} & {0.62} & {0.807} & {0.711} & \textbf{0.823} & {0.662} \tabularnewline
{GunPoint} & {0.993} & {0.907} & \textbf{1} & \textbf{1} & \textbf{1} & {0.993} \tabularnewline
{Ham} & \textbf{0.695} & {0.467} & {0.686} & {0.667} & {0.667} & {0.571} \tabularnewline
{HandOutlines} & {0.922} & {0.881} & \textbf{0.932} & {0.903} & \textbf{0.932} & {0.889} \tabularnewline
{Haptics} & {0.455} & {0.377} & \textbf{0.523} & {0.461} & {0.519} & {0.393} \tabularnewline
{Herring} & {0.578} & {0.531} & {0.672} & {0.547} & \textbf{0.688} & {0.578} \tabularnewline
{InlineSkate} & {0.447} & {0.384} & {0.373} & \textbf{0.516} & {0.5} & {0.46} \tabularnewline
{InsectWingbeatSound} & {0.623} & {0.355} & {0.627} & {0.523} & \textbf{0.655} & {0.595} \tabularnewline
{ItalyPowerDemand} & {0.925} & {0.95} & {0.948} & {0.909} & \textbf{0.963} & {0.962} \tabularnewline
{LargeKitchenAppliances} & {0.848} & {0.795} & {0.859} & {0.765} & \textbf{0.864} & {0.811} \tabularnewline
{Lightning2} & \textbf{0.918} & {0.869} & {0.738} & {0.836} & {0.82} & {0.885} \tabularnewline
{Lightning7} & \textbf{0.795} & {0.726} & {0.726} & {0.685} & {0.74} & {0.767} \tabularnewline
{Mallat} & \textbf{0.964} & {0.934} & \textbf{0.964} & {0.938} & {0.962} & {0.94} \tabularnewline
{Meat} & \textbf{0.95} & {0.933} & {0.85} & {0.9} & {0.933} & {0.933} \tabularnewline
{MedicalImages} & \textbf{0.784} & {0.737} & {0.67} & {0.718} & {0.778} & {0.742} \tabularnewline
{MiddlePhalanxOutlineCorrect} & {0.814} & {0.698} & {0.794} & {0.78} & \textbf{0.832} & {0.784} \tabularnewline
{MiddlePhalanxOutlineAgeGroup} & \textbf{0.656} & {0.5} & {0.643} & {0.545} & {0.597} & {0.558} \tabularnewline
{MiddlePhalanxTW} & \textbf{0.61} & {0.506} & {0.519} & {0.545} & {0.571} & {0.513} \tabularnewline
{MoteStrain} & {0.871} & {0.835} & {0.897} & {0.879} & \textbf{0.933} & {0.883} \tabularnewline
{NonInvasiveFetalECGThorax1} & {0.91} & {0.79} & \textbf{0.95} & {0.838} & {0.93} & {0.846} \tabularnewline
{NonInvasiveFetalECGThorax2} & {0.927} & {0.865} & \textbf{0.951} & {0.901} & {0.945} & {0.913} \tabularnewline
{OliveOil} & \textbf{0.9} & {0.833} & \textbf{0.9} & {0.867} & \textbf{0.9} & {0.867} \tabularnewline
{OSULeaf} & {0.831} & {0.591} & {0.967} & {0.955} & \textbf{0.979} & {0.806} \tabularnewline
{PhalangesOutlinesCorrect} & {0.801} & {0.728} & {0.763} & {0.772} & \textbf{0.807} & {0.773} \tabularnewline
{Phoneme} & {0.289} & {0.228} & {0.321} & {0.265} & \textbf{0.382} & {0.305} \tabularnewline
{Plane} & {0.99} & \textbf{1} & \textbf{1} & \textbf{1} & \textbf{1} & \textbf{1} \tabularnewline
{ProximalPhalanxOutlineCorrect} & {0.859} & {0.784} & \textbf{0.883} & {0.849} & {0.88} & {0.808} \tabularnewline
{ProximalPhalanxOutlineAgeGroup} & {0.854} & {0.805} & {0.844} & {0.834} & \textbf{0.859} & {0.805} \tabularnewline
{ProximalPhalanxTW} & \textbf{0.824} & {0.761} & {0.805} & {0.8} & {0.815} & {0.766} \tabularnewline
{RefrigerationDevices} & {0.517} & {0.464} & \textbf{0.581} & {0.499} & {0.557} & {0.437} \tabularnewline
{ScreenType} & {0.413} & {0.397} & {0.52} & {0.464} & \textbf{0.589} & {0.445} \tabularnewline
{ShapeletSim} & {0.817} & {0.65} & {0.956} & \textbf{1} & \textbf{1} & {0.817} \tabularnewline
{ShapesAll} & {0.875} & {0.768} & {0.842} & \textbf{0.908} & {0.905} & {0.867} \tabularnewline
{SmallKitchenAppliances} & {0.715} & {0.643} & {0.792} & {0.725} & \textbf{0.853} & {0.696} \tabularnewline
{SonyAIBORobotSurface1} & \textbf{0.897} & {0.725} & {0.844} & {0.632} & {0.765} & {0.704} \tabularnewline
{SonyAIBORobotSurface2} & \textbf{0.934} & {0.831} & \textbf{0.934} & {0.859} & {0.928} & {0.878} \tabularnewline
{StarLightCurves} & {0.965} & {0.907} & {0.979} & {0.978} & \textbf{0.982} & {0.926} \tabularnewline
{Strawberry} & {0.946} & {0.941} & {0.962} & \textbf{0.976} & {0.97} & {0.946} \tabularnewline
{SwedishLeaf} & {0.931} & {0.792} & {0.928} & {0.922} & \textbf{0.954} & {0.915} \tabularnewline
{Symbols} & {0.965} & {0.95} & {0.882} & {0.967} & \textbf{0.974} & {0.96} \tabularnewline
{SyntheticControl} & {0.983} & {0.993} & {0.983} & {0.967} & \textbf{0.997} & {0.99} \tabularnewline
{ToeSegmentation1} & {0.952} & {0.772} & {0.965} & {0.939} & \textbf{0.982} & {0.829} \tabularnewline
{ToeSegmentation2} & {0.885} & {0.838} & {0.908} & \textbf{0.962} & {0.954} & {0.892} \tabularnewline
{Trace} & \textbf{1} & \textbf{1} & \textbf{1} & \textbf{1} & \textbf{1} & {0.99} \tabularnewline
{TwoLeadECG} & \textbf{0.997} & {0.905} & \textbf{0.997} & {0.981} & {0.996} & {0.971} \tabularnewline
{TwoPatterns} & \textbf{1} & \textbf{1} & {0.955} & {0.993} & \textbf{1} & \textbf{1} \tabularnewline
{UWaveGestureLibraryX} & {0.811} & {0.728} & {0.803} & {0.762} & \textbf{0.84} & {0.805} \tabularnewline
{UWaveGestureLibraryY} & {0.735} & {0.634} & {0.73} & {0.685} & \textbf{0.765} & {0.726} \tabularnewline
{UWaveGestureLibraryZ} & {0.759} & {0.658} & {0.748} & {0.695} & \textbf{0.783} & {0.724} \tabularnewline
{UWaveGestureLibraryAll} & {0.941} & {0.892} & {0.942} & {0.939} & \textbf{0.968} & \textbf{0.968} \tabularnewline
{Wafer} & {0.993} & {0.98} & \textbf{1} & {0.995} & {0.999} & {0.997} \tabularnewline
{Wine} & \textbf{0.87} & {0.574} & {0.796} & {0.741} & {0.778} & {0.574} \tabularnewline
{WordSynonyms} & {0.704} & {0.649} & {0.571} & {0.638} & {0.738} & \textbf{0.779} \tabularnewline
{Worms} & {0.714} & {0.584} & \textbf{0.74} & {0.558} & {0.558} & {0.662} \tabularnewline
{WormsTwoClass} & {0.818} & {0.623} & \textbf{0.831} & \textbf{0.831} & {0.779} & {0.688} \tabularnewline
{Yoga} & {0.878} & {0.837} & {0.818} & \textbf{0.918} & \textbf{0.918} & {0.879} \tabularnewline
\bottomrule
\end{longtable}
\endgroup

\newpage{}

\begingroup
\scriptsize
\begin{longtable}[c]{lcccc}
\caption{\label{app_table:comparison_to_others_deep}Accuracy scores of the combined version of our method compared with
those of ResNet (supervised), TimeNet and RWS (unsupervised), when available. Bold scores indicate
the best performing method. ``X''s indicate that a score was reported
in the original paper, but was either obtained using transferability
or on a reversed train~/~test split of the dataset, thus not comparable
to other results for this dataset.}
\tabularnewline
\toprule
\multirow{3}{*}[-\dimexpr \aboverulesep + \belowrulesep + \cmidrulewidth]{{Dataset}} & {Unsupervised} & {Supervised} & \multicolumn{2}{c}{{Unsupervised}} \tabularnewline
\cmidrule(lr){2-2} \cmidrule(lr){3-3} \cmidrule(lr){4-5}
& {Ours} & \multirow{2}{*}[-0.5\dimexpr \aboverulesep + \belowrulesep + \cmidrulewidth]{{ResNet}} & \multirow{2}{*}[-0.5\dimexpr \aboverulesep + \belowrulesep + \cmidrulewidth]{{TimeNet}} & \multirow{2}{*}[-0.5\dimexpr \aboverulesep + \belowrulesep + \cmidrulewidth]{{RWS}} \tabularnewline
\cmidrule(lr){2-2}
& {Combined} & & & \tabularnewline
\midrule
\endhead
\bottomrule
\endlastfoot
{Adiac} & {0.716} & \textbf{0.831} & {0.565} & {-} \tabularnewline
{ArrowHead} & {0.829} & \textbf{0.84} & {-} & {-} \tabularnewline
{Beef} & {0.7} & \textbf{0.767} & {-} & {0.733} \tabularnewline
{BeetleFly} & \textbf{0.9} & {0.85} & {-} & {-} \tabularnewline
{BirdChicken} & {0.8} & \textbf{0.95} & {-} & {-} \tabularnewline
{Car} & {0.817} & \textbf{0.917} & {-} & {-} \tabularnewline
{CBF} & \textbf{0.994} & {0.989} & {-} & {-} \tabularnewline
{ChlorineConcentration} & {0.782} & \textbf{0.835} & {0.723} & {0.572} \tabularnewline
{CinCECGTorso} & {0.74} & \textbf{0.838} & {-} & {-} \tabularnewline
{Coffee} & \textbf{1} & \textbf{1} & {-} & {-} \tabularnewline
{Computers} & {0.628} & \textbf{0.816} & {-} & {-} \tabularnewline
{CricketX} & {0.777} & \textbf{0.79} & {0.659} & {-} \tabularnewline
{CricketY} & {0.767} & \textbf{0.805} & {X} & {-} \tabularnewline
{CricketZ} & {0.764} & \textbf{0.831} & {X} & {-} \tabularnewline
{DiatomSizeReduction} & \textbf{0.993} & {0.301} & {-} & {-} \tabularnewline
{DistalPhalanxOutlineCorrect} & \textbf{0.768} & {X} & {X} & {-} \tabularnewline
{DistalPhalanxOutlineAgeGroup} & \textbf{0.734} & {X} & {X} & {-} \tabularnewline
{DistalPhalanxTW} & \textbf{0.676} & {X} & {X} & {X} \tabularnewline
{Earthquakes} & \textbf{0.748} & {X} & {-} & {-} \tabularnewline
{ECG200} & \textbf{0.9} & {0.87} & {-} & {-} \tabularnewline
{ECG5000} & \textbf{0.936} & {0.935} & {0.934} & {0.933} \tabularnewline
{ECGFiveDays} & \textbf{1} & {0.99} & {X} & {-} \tabularnewline
{ElectricDevices} & {0.732} & \textbf{0.735} & {0.665} & {-} \tabularnewline
{FaceAll} & {0.802} & \textbf{0.855} & {-} & {-} \tabularnewline
{FaceFour} & {0.875} & \textbf{0.955} & {-} & {-} \tabularnewline
{FacesUCR} & {0.918} & \textbf{0.955} & {-} & {-} \tabularnewline
{FiftyWords} & \textbf{0.78} & {0.732} & {-} & {-} \tabularnewline
{Fish} & {0.88} & \textbf{0.977} & {-} & {-} \tabularnewline
{FordA} & \textbf{0.935} & {X} & {X} & {-} \tabularnewline
{FordB} & \textbf{0.81} & {X} & {X} & {X} \tabularnewline
{GunPoint} & \textbf{0.993} & \textbf{0.993} & {-} & {-} \tabularnewline
{Ham} & {0.695} & \textbf{0.8} & {-} & {-} \tabularnewline
{HandOutlines} & \textbf{0.922} & {X} & {-} & {X} \tabularnewline
{Haptics} & {0.455} & \textbf{0.516} & {-} & {-} \tabularnewline
{Herring} & {0.578} & \textbf{0.641} & {-} & {-} \tabularnewline
{InlineSkate} & \textbf{0.447} & {0.378} & {-} & {-} \tabularnewline
{InsectWingbeatSound} & \textbf{0.623} & {0.506} & {-} & {0.619} \tabularnewline
{ItalyPowerDemand} & {0.925} & {0.959} & {-} & \textbf{0.969} \tabularnewline
{LargeKitchenAppliances} & {0.848} & \textbf{0.904} & {-} & {0.792} \tabularnewline
{Lightning2} & \textbf{0.918} & {0.77} & {-} & {-} \tabularnewline
{Lightning7} & {0.795} & \textbf{0.863} & {-} & {-} \tabularnewline
{Mallat} & {0.964} & \textbf{0.966} & {-} & {0.937} \tabularnewline
{Meat} & {0.95} & \textbf{0.983} & {-} & {-} \tabularnewline
{MedicalImages} & \textbf{0.784} & {0.762} & {0.753} & {-} \tabularnewline
{MiddlePhalanxOutlineCorrect} & \textbf{0.814} & {X} & {X} & {X} \tabularnewline
{MiddlePhalanxOutlineAgeGroup} & \textbf{0.656} & {X} & {X} & {-} \tabularnewline
{MiddlePhalanxTW} & \textbf{0.61} & {X} & {X} & {-} \tabularnewline
{MoteStrain} & {0.871} & \textbf{0.924} & {-} & {-} \tabularnewline
{NonInvasiveFetalECGThorax1} & {0.91} & \textbf{0.946} & {-} & {0.907} \tabularnewline
{NonInvasiveFetalECGThorax2} & {0.927} & \textbf{0.944} & {-} & {-} \tabularnewline
{OliveOil} & \textbf{0.9} & {0.867} & {-} & {-} \tabularnewline
{OSULeaf} & {0.831} & \textbf{0.979} & {-} & {-} \tabularnewline
{PhalangesOutlinesCorrect} & {0.801} & \textbf{0.857} & {0.772} & {-} \tabularnewline
{Phoneme} & {0.289} & \textbf{0.333} & {-} & {-} \tabularnewline
{Plane} & {0.99} & \textbf{1} & {-} & {-} \tabularnewline
{ProximalPhalanxOutlineCorrect} & {0.859} & \textbf{0.914} & {X} & {0.711} \tabularnewline
{ProximalPhalanxOutlineAgeGroup} & \textbf{0.854} & {0.839} & {X} & {X} \tabularnewline
{ProximalPhalanxTW} & \textbf{0.824} & {X} & {X} & {-} \tabularnewline
{RefrigerationDevices} & \textbf{0.517} & \textbf{0.517} & {-} & {-} \tabularnewline
{ScreenType} & {0.413} & \textbf{0.632} & {-} & {-} \tabularnewline
{ShapeletSim} & {0.817} & \textbf{1} & {-} & {-} \tabularnewline
{ShapesAll} & {0.875} & \textbf{0.917} & {-} & {-} \tabularnewline
{SmallKitchenAppliances} & {0.715} & \textbf{0.789} & {-} & {-} \tabularnewline
{SonyAIBORobotSurface1} & {0.897} & \textbf{0.968} & {-} & {-} \tabularnewline
{SonyAIBORobotSurface2} & {0.934} & \textbf{0.986} & {-} & {-} \tabularnewline
{StarLightCurves} & {0.965} & \textbf{0.972} & {-} & {-} \tabularnewline
{Strawberry} & \textbf{0.946} & {X} & {X} & {-} \tabularnewline
{SwedishLeaf} & {0.931} & \textbf{0.955} & {0.901} & {-} \tabularnewline
{Symbols} & \textbf{0.965} & {0.927} & {-} & {-} \tabularnewline
{SyntheticControl} & {0.983} & \textbf{1} & {0.983} & {-} \tabularnewline
{ToeSegmentation1} & {0.952} & \textbf{0.969} & {-} & {-} \tabularnewline
{ToeSegmentation2} & {0.885} & \textbf{0.915} & {-} & {-} \tabularnewline
{Trace} & \textbf{1} & \textbf{1} & {-} & {-} \tabularnewline
{TwoLeadECG} & {0.997} & \textbf{1} & {-} & {-} \tabularnewline
{TwoPatterns} & \textbf{1} & \textbf{1} & {0.999} & {0.999} \tabularnewline
{UWaveGestureLibraryX} & \textbf{0.811} & {0.78} & {-} & {-} \tabularnewline
{UWaveGestureLibraryY} & \textbf{0.735} & {0.675} & {-} & {-} \tabularnewline
{UWaveGestureLibraryZ} & \textbf{0.759} & {0.75} & {-} & {-} \tabularnewline
{UWaveGestureLibraryAll} & \textbf{0.941} & {0.862} & {-} & {-} \tabularnewline
{Wafer} & {0.993} & \textbf{0.998} & {0.994} & {0.993} \tabularnewline
{Wine} & \textbf{0.87} & {0.611} & {-} & {-} \tabularnewline
{WordSynonyms} & \textbf{0.704} & {0.625} & {-} & {-} \tabularnewline
{Worms} & \textbf{0.714} & {X} & {-} & {-} \tabularnewline
{WormsTwoClass} & \textbf{0.818} & {X} & {-} & {-} \tabularnewline
{Yoga} & \textbf{0.878} & {0.857} & {0.866} & {-} \tabularnewline
\end{longtable}
\endgroup

\newpage

\begin{table}[h!]
\caption{\label{app_table:our_results_ucr_newest}Accuracy scores of variants of our method and of DTW on the remaining 43 UCR datasets, except DodgerLoopDay, DodgerLoopGame and DodgerLoopWeekend which contain missing values. Bold scores indicate the best performing method.}
\begin{centering}
\scriptsize
\begin{tabular}{lcccccccc}
\toprule
\multirow{3}{*}[-\dimexpr \aboverulesep + \belowrulesep + \cmidrulewidth]{{Dataset}} & \multicolumn{8}{c}{{Unsupervised}} \tabularnewline
\cmidrule(lr){2-9}
& \multicolumn{7}{c}{{Ours}} & {\multirow{2}{*}[-0.5\dimexpr \aboverulesep + \belowrulesep + \cmidrulewidth]{{DTW}}} \tabularnewline
\cmidrule(lr){2-8}
& {$K=1$} & {$K=2$} & {$K=5$} & {$K=10$} & {Combined} & {Combined ($1$-NN)} & {FordA}& \tabularnewline\midrule
{ACSF1} & \textbf{0.91} & {0.87} & {0.86} & {0.9} & {0.81} & {0.85} & {0.73} & {0.64} \tabularnewline
{AllGestureWiimoteX} & {0.721} & {0.746} & {0.747} & {0.763} & \textbf{0.779} & {0.736} & {0.693} & {0.716} \tabularnewline
{AllGestureWiimoteY} & {0.741} & {0.744} & {0.759} & {0.726} & \textbf{0.793} & {0.756} & {0.713} & {0.729} \tabularnewline
{AllGestureWiimoteZ} & {0.687} & {0.697} & {0.691} & {0.723} & \textbf{0.763} & {0.716} & {0.71} & {0.643} \tabularnewline
{BME} & \textbf{0.993} & \textbf{0.993} & {0.987} & \textbf{0.993} & \textbf{0.993} & {0.947} & {0.96} & {0.9} \tabularnewline
{Chinatown} & {0.951} & {0.951} & {0.942} & {0.951} & \textbf{0.962} & {0.936} & \textbf{0.962} & {0.957} \tabularnewline
{Crop} & {0.728} & {0.726} & {0.728} & {0.722} & \textbf{0.746} & {0.695} & {0.727} & {0.665} \tabularnewline
{EOGHorizontalSignal} & {0.552} & {0.566} & {0.536} & \textbf{0.605} & {0.588} & {0.522} & {0.47} & {0.503} \tabularnewline
{EOGVerticalSignal} & {0.398} & {0.414} & {0.431} & {0.434} & \textbf{0.489} & {0.431} & {0.439} & {0.448} \tabularnewline
{EthanolLevel} & {0.418} & {0.34} & {0.316} & {0.382} & {0.392} & {0.274} & \textbf{0.558} & {0.276} \tabularnewline
{FreezerRegularTrain} & {0.986} & {0.988} & {0.979} & {0.956} & {0.955} & {0.963} & \textbf{0.992} & {0.899} \tabularnewline
{FreezerSmallTrain} & {0.967} & {0.956} & {0.906} & \textbf{0.933} & {0.928} & {0.872} & {0.862} & {0.753} \tabularnewline
{Fungi} & \textbf{1} & \textbf{1} & \textbf{1} & \textbf{1} & \textbf{1} & \textbf{1} & {0.925} & {0.839} \tabularnewline
{GestureMidAirD1} & \textbf{0.638} & {0.577} & {0.592} & {0.608} & {0.615} & {0.546} & {0.608} & {0.569} \tabularnewline
{GestureMidAirD2} & {0.508} & {0.515} & {0.523} & \textbf{0.546} & {0.508} & {0.415} & {0.538} & {0.608} \tabularnewline
{GestureMidAirD3} & {0.269} & \textbf{0.331} & {0.308} & {0.285} & \textbf{0.331} & {0.246} & {0.292} & {0.323} \tabularnewline
{GesturePebbleZ1} & {0.826} & {0.843} & {0.913} & {0.919} & \textbf{0.936} & {0.814} & {0.547} & {0.791} \tabularnewline
{GesturePebbleZ2} & {0.861} & {0.873} & {0.88} & \textbf{0.899} & {0.88} & {0.791} & {0.538} & {0.671} \tabularnewline
{GunPointAgeSpan} & {0.984} & {0.984} & \textbf{0.994} & \textbf{0.994} & {0.987} & {0.991} & {0.987} & {0.918} \tabularnewline
{GunPointMaleVersusFemale} & \textbf{1} & \textbf{1} & \textbf{1} & {0.997} & \textbf{1} & {0.994} & \textbf{1} & {0.997} \tabularnewline
{GunPointOldVersusYoung} & \textbf{1} & \textbf{1} & \textbf{1} & \textbf{1} & \textbf{1} & \textbf{1} & \textbf{1} & {0.838} \tabularnewline
{HouseTwenty} & \textbf{0.95} & {0.933} & {0.916} & {0.933} & \textbf{0.95} & {0.924} & {0.882} & {0.924} \tabularnewline
{InsectEPGRegularTrain} & \textbf{1} & \textbf{1} & \textbf{1} & \textbf{1} & \textbf{1} & \textbf{1} & \textbf{1} & {0.872} \tabularnewline
{InsectEPGSmallTrain} & \textbf{1} & \textbf{1} & \textbf{1} & \textbf{1} & \textbf{1} & \textbf{1} & \textbf{1} & {0.735} \tabularnewline
{MelbournePedestrian} & {0.949} & {0.946} & {0.943} & {0.944} & \textbf{0.951} & {0.914} & {0.947} & {0.791} \tabularnewline
{MixedShapesRegularTrain} & {0.916} & {0.906} & {0.904} & {0.905} & \textbf{0.927} & {0.898} & {0.898} & {0.842} \tabularnewline
{MixedShapesSmallTrain} & {0.864} & {0.857} & {0.871} & {0.86} & \textbf{0.877} & {0.829} & {0.861} & {0.78} \tabularnewline
{PickupGestureWiimoteZ} & {0.72} & \textbf{0.8} & {0.78} & {0.74} & {0.78} & {0.72} & {0.74} & {0.66} \tabularnewline
{PigAirwayPressure} & {0.385} & {0.452} & \textbf{0.51} & \textbf{0.51} & {0.486} & {0.332} & {0.317} & {0.106} \tabularnewline
{PigArtPressure} & {0.88} & {0.933} & \textbf{0.942} & {0.928} & {0.933} & {0.861} & {0.591} & {0.245} \tabularnewline
{PigCVP} & {0.404} & {0.548} & {0.62} & \textbf{0.788} & {0.712} & {0.385} & {0.534} & {0.154} \tabularnewline
{PLAID} & {0.533} & {0.549} & {0.574} & {0.555} & {0.559} & {0.696} & {0.493} & \textbf{0.84} \tabularnewline
{PowerCons} & \textbf{0.961} & {0.939} & {0.9} & {0.9} & {0.928} & {0.894} & {0.933} & {0.878} \tabularnewline
{Rock} & {0.62} & {0.62} & {0.58} & {0.58} & \textbf{0.68} & {0.5} & {0.54} & {0.6} \tabularnewline
{SemgHandGenderCh2} & {0.845} & {0.852} & {0.873} & {0.89} & \textbf{0.902} & {0.863} & {0.84} & {0.802} \tabularnewline
{SemgHandMovementCh2} & {0.711} & {0.649} & {0.7} & \textbf{0.789} & {0.784} & {0.709} & {0.516} & {0.584} \tabularnewline
{SemgHandSubjectCh2} & {0.767} & {0.816} & {0.851} & {0.853} & \textbf{0.876} & {0.72} & {0.591} & {0.727} \tabularnewline
{ShakeGestureWiimoteZ} & {0.92} & {0.92} & \textbf{0.94} & {0.92} & \textbf{0.94} & {0.86} & {0.9} & {0.86} \tabularnewline
{SmoothSubspace} & {0.933} & \textbf{0.96} & {0.94} & \textbf{0.96} & {0.953} & {0.833} & {0.94} & {0.827} \tabularnewline
{UMD} & {0.979} & {0.986} & \textbf{0.993} & \textbf{0.993} & \textbf{0.993} & {0.958} & {0.986} & \textbf{0.993} \tabularnewline
\bottomrule
\end{tabular}
\end{centering}
\end{table}

\newpage

\section{Multivariate Time Series}
\label{app:multivariate}

Full results corresponding to the UEA archive datasets for our method as well as the ones of DTW\textsubscript{D} as reported by \citet{Bagnall2018} are presented in Table~\ref{app_table:multivariate}, for the unique train / test split provided in the archive.

\begin{table}[h!]
\caption{Accuracy scores of variants of our method on all UEA datasets, compared to DTW\textsubscript{D}. Bold scores indicate the best performing method.\label{app_table:multivariate}}
\begin{centering}
\scriptsize
\begin{tabular}{lccccc}
\toprule
\multirow{3}{*}[-\dimexpr \aboverulesep + \belowrulesep + \cmidrulewidth]{{Dataset}} & \multicolumn{5}{c}{{Unsupervised}} \tabularnewline
\cmidrule(lr){2-6}
& \multicolumn{4}{c}{{Ours}} & \multicolumn{1}{c}{{\multirow{2}{*}[-0.5\dimexpr \aboverulesep + \belowrulesep + \cmidrulewidth]{{{DTW}\textsubscript{D}}}}} \tabularnewline
\cmidrule(lr){2-5}
& {$K=5$} & {$K=10$} & {$K=20$} & {Combined} & \tabularnewline
\midrule
{ArticularyWordRecognition} & {0.967} & {0.973} & {0.943} & \textbf{0.987} & \textbf{0.987} \tabularnewline
{AtrialFibrillation} & \textbf{0.2} & {0.067} & {0.133} & {0.133} & \textbf{0.2} \tabularnewline
{BasicMotions} & \textbf{1} & \textbf{1} & \textbf{1} & \textbf{1} & {0.975} \tabularnewline
{CharacterTrajectories} & {0.986} & {0.99} & {0.993} & \textbf{0.994} & {0.989} \tabularnewline
{Cricket} & {0.958} & {0.972} & {0.972} & {0.986} & \textbf{1} \tabularnewline
{DuckDuckGeese} & {0.6} & \textbf{0.675} & {0.65} & \textbf{0.675} & {0.6} \tabularnewline
{EigenWorms} & {0.87} & {0.802} & {0.84} & \textbf{0.878} & {0.618} \tabularnewline
{Epilepsy} & \textbf{0.971} & \textbf{0.971} & \textbf{0.971} & {0.957} & {0.964} \tabularnewline
{Ering} & \textbf{0.133} & \textbf{0.133} & \textbf{0.133} & \textbf{0.133} & \textbf{0.133} \tabularnewline
{EthanolConcentration} & {0.289} & {0.251} & {0.205} & {0.236} & \textbf{0.323} \tabularnewline
{FaceDetection} & {0.522} & {0.525} & {0.513} & {0.528} & \textbf{0.529} \tabularnewline
{FingerMovements} & {0.55} & {0.49} & \textbf{0.58} & {0.54} & {0.53} \tabularnewline
{HandMovementDirection} & {0.311} & {0.297} & \textbf{0.351} & {0.27} & {0.231} \tabularnewline
{Handwriting} & {0.447} & {0.464} & {0.451} & \textbf{0.533} & {0.286} \tabularnewline
{Heartbeat} & \textbf{0.756} & {0.732} & {0.741} & {0.737} & {0.717} \tabularnewline
{InsectWingbeat} & {0.159} & {0.158} & {0.156} & \textbf{0.16} & {-} \tabularnewline
{JapaneseVowels} & {0.984} & {0.986} & \textbf{0.989} & \textbf{0.989} & {0.949} \tabularnewline
{Libras} & {0.878} & \textbf{0.883} & \textbf{0.883} & {0.867} & {0.87} \tabularnewline
{LSST} & {0.535} & {0.552} & {0.509} & \textbf{0.558} & {0.551} \tabularnewline
{MotorImagery} & {0.53} & {0.54} & \textbf{0.58} & {0.54} & {0.5} \tabularnewline
{NATOPS} & {0.933} & {0.917} & {0.917} & \textbf{0.944} & {0.883} \tabularnewline
{PEMS-SF} & {0.636} & {0.671} & {0.676} & \textbf{0.688} & {0.711} \tabularnewline
{PenDigits} & \textbf{0.985} & {0.979} & {0.981} & {0.983} & {0.977} \tabularnewline
{Phoneme} & {0.216} & {0.214} & {0.222} & \textbf{0.246} & {0.151} \tabularnewline
{RacketSports} & {0.776} & {0.836} & {0.855} & \textbf{0.862} & {0.803} \tabularnewline
{SelfRegulationSCP1} & {0.795} & {0.826} & {0.843} & \textbf{0.846} & {0.775} \tabularnewline
{SelfRegulationSCP2} & {0.55} & {0.539} & {0.539} & \textbf{0.556} & {0.539} \tabularnewline
{SpokenArabicDigits} & {0.908} & {0.894} & {0.905} & {0.956} & \textbf{0.963} \tabularnewline
{StandWalkJump} & {0.333} & \textbf{0.4} & {0.333} & \textbf{0.4} & {0.2} \tabularnewline
{UWaveGestureLibrary} & {0.884} & {0.869} & {0.875} & {0.884} & \textbf{0.903} \tabularnewline
\bottomrule
\end{tabular}
\end{centering}
\end{table}

\newpage

\section{Discussion of the Choice of Encoder}
\label{app:EncoderDiscussion}

One of the aims of this work is to propose a representation learning method for time series that is scalable.
For this reason, and as explained in Section~\ref{sec:EncoderArchitecture}, we did not consider using an LSTM
as encoder $\ff$. Nonetheless, we experimented with such an encoder on a small set of UCR datasets in order to
get an indication of its performance versus the proposed encoder in this paper. We use the same optimization
hyperparameters as those used to train the causal CNN encoder and choose a two-layer LSTM with hidden size $256$
in order to compare both networks with similar computational time and memory usage. Corresponding results are
compiled in Table~\ref{app_table:lstm}.

We observe on this restricted set of experiments that not only the proposed encoder outperforms the LSTM
encoder, but it does so by a large margin. This indicates that the proposed causal CNN encoder is more adapted
to the considered task and training method.

\begin{table}[h!]
\caption{\label{app_table:lstm}Results of our training method combined with the proposed causal CNN encoder on one hand, and with an LSTM encoder on the other hand ($K = 5$). Bold scores indicate the best performing method.}
\begin{centering}
\scriptsize
\begin{tabular}{lcc}
\toprule
{Dataset} & {Causal CNN} & {LSTM} \tabularnewline
\midrule
{Adiac} & \textbf{0.703} & {0.269} \tabularnewline
{Computers} & \textbf{0.676} & {0.492} \tabularnewline
{CricketX} & \textbf{0.7} & {0.136}\tabularnewline
{DistalPhalanxTW} & \textbf{0.662} & \textbf{0.662} \tabularnewline
{Earthquakes} & \textbf{0.748} & \textbf{0.748} \tabularnewline
{HandOutlines} & \textbf{0.908} & {0.646} \tabularnewline
{NonInvasiveFetalECGThorax1} & \textbf{0.904} & {0.169} \tabularnewline
{PhalangesOutlinesCorrect} & \textbf{0.795} & {0.613} \tabularnewline
{RefrigerationDevices} & \textbf{0.547} & {0.411} \tabularnewline
{UWaveGestureLibraryX} & \textbf{0.806} & {0.357} \tabularnewline
{Wafer} & \textbf{0.993} & {0.896} \tabularnewline
\bottomrule
\end{tabular}
\end{centering}
\end{table}

\end{document}